\documentclass[lettersize,journal]{IEEEtran}
\usepackage{amsmath,amsfonts}
\usepackage{algorithmic}
\usepackage{algorithm}
\usepackage{array}
\usepackage[caption=false,font=normalsize,labelfont=sf,textfont=sf]{subfig}
\usepackage{textcomp}
\usepackage{stfloats}
\usepackage{url}
\usepackage{verbatim}
\usepackage{graphicx}
\usepackage{cite}
\usepackage{xcolor}
\newcommand{\revise}[1]{#1}                       
\newcommand{\rtwo}[1]{\textcolor{black}{#1}}        
\newcommand{\ci}[1]{{\scriptsize[#1]}}

\usepackage{listings}
\definecolor{codegray}{rgb}{0.95,0.95,0.95}
\definecolor{keywordblue}{rgb}{0.0,0.0,0.6}
\definecolor{stringred}{rgb}{0.6,0.0,0.0}
\definecolor{importgreen}{rgb}{0.0,0.5,0.0}
\definecolor{textblack}{rgb}{0.0,0.0,0.0}
\lstdefinestyle{json}{
  language={},
  basicstyle=\ttfamily\scriptsize,
  breaklines=true,
  breakatwhitespace=false,
  columns=fullflexible,
  keepspaces=true,
  postbreak=\mbox{\textcolor{red}{$\hookrightarrow$}\space}
}

\usepackage{tabularx,array}

\usepackage{booktabs}
\usepackage{multirow,makecell}

\usepackage[colorlinks=true,linkcolor=blue,citecolor=blue,urlcolor=blue]{hyperref}
\hypersetup{pdfauthor={Haoyang Xie and Feng Ju}}

\begin{document}

\title{Text-to-CadQuery: Automating CAD Model Generation from Text with Scalable Large Language Models}

\author{Haoyang Xie, Feng Ju
\thanks{Haoyang Xie and Feng Ju are with the School of Computing and Augmented Intelligence, Arizona State University, Tempe, AZ 85281 USA (e-mail: hxie40@asu.edu; fengju@asu.edu).}
\thanks{Corresponding author: Feng Ju.}}
\markboth{IEEE Transactions on Automation Science and Engineering}%
{Xie and Ju: Text-to-CadQuery: Automating CAD Model Generation from Text}

\IEEEaftertitletext{\vspace{-1.5\baselineskip}}
\maketitle

\begin{abstract}
Computer-aided design (CAD) is fundamental to modern engineering and manufacturing, but creating CAD models still requires expert knowledge and specialized software. Recent advances in large language models (LLMs) open up the possibility of generative CAD, where natural language is directly translated into parametric 3D models. However, most existing methods generate task-specific command sequences\revise{, a custom token format that pretrained models have not seen and that needs dataset-specific conversion code to produce a 3D model and prior methods built on them train task-specific models from scratch.} To tackle this issue, we propose generating \revise{CadQuery, a Python-based CAD scripting language,} directly from text. \revise{We} augment the Text2CAD dataset with 170,000 CadQuery annotations\revise{, obtained by deterministic translation and verified geometrically against the source CAD models}. \revise{We fine-tune four sizes of one model family, Qwen2.5 (0.5B to 7B), under identical settings and observe consistent improvements with size. On held-out models, our best model outperforms the Text2CAD baseline by a wide margin: it raises F1 from 0.34 to 0.69 and volumetric IoU from 0.38 to 0.74 and raises the share of outputs judged by a vision-language model to match the reference from 28.0\% to 65.4\%.} \revise{Compared with a command-sequence baseline on the same backbone, CadQuery reaches the same shape accuracy, is slightly more accurate in absolute dimensions, transfers better \rtwo{on a small external benchmark}, and yields readable, editable programs that export directly to STEP without dataset-specific conversion code.} Project page: \url{https://github.com/Text-to-CadQuery/Text-to-CadQuery}.
\end{abstract}

\textbf{
\textit{Note to Practitioners}---This work addresses the difficulty of creating CAD models with specialized software and expert knowledge. The proposed system allows users to describe a part in natural language and automatically generates CadQuery scripts for parametric modeling. By running the generated scripts, users can directly obtain 3D models and export them to common industrial formats such as STL, STEP, and B-rep. This makes the output inspectable, editable, and easy to integrate into existing design automation, prototyping, and manufacturing workflows. The approach can reduce early-stage modeling effort and lower the barrier for non-expert users. In practical use, the generated models should be checked for executability and geometric validity before being used in downstream manufacturing or production workflows.
}


\section{Introduction}\label{sec:introduction}
\IEEEPARstart{C}{omputer}-aided design (CAD) plays a central role in engineering, architecture, construction, and 3D printing. It is a core tool for creating precise digital models that support analysis, simulation, and fabrication~\cite{hunde2022future,fathizadan2022dynamic, xie2025transformer}. Beyond these well-known domains, CAD has become indispensable in industrial design, aerospace engineering, and robotics manufacturing, where digital models are not only used for geometric modeling but also serve as the foundation for structural simulation, assembly verification, and manufacturability analysis. In these contexts, CAD acts as the connective tissue linking design, simulation, and production, and its efficiency directly impacts the speed and quality of the overall design-to-manufacturing pipeline. 

Despite its importance, creating CAD models remains a highly complex process. Designers must master thousands of commands, constraints, and software-specific conventions, making the learning curve steep even for technically trained users. Moreover, traditional CAD tools remain inaccessible to many potential contributors outside the core group of trained designers. During early-stage ideation, when ideas from engineers, product managers, or domain specialists could be valuable, the steep technical requirements of CAD software often prevent them from actively participating. As a result, design concepts that originate from diverse perspectives are either lost or delayed, slowing innovation and limiting the inclusiveness of the overall design process. 

The integration of Artificial Intelligence~(AI) with CAD systems offers a promising direction to address these challenges~\cite{xie2026agent}. By lowering entry barriers and automating parts of the modeling process, AI-driven approaches can make CAD more accessible to non-experts while also improving the efficiency of expert designers. With the recent rise of foundation models, including large language and multimodal models, it has become feasible to enable more intuitive interaction with CAD systems through natural language, images, sketches, and point clouds~\cite{zooZooModel}.

Industry systems such as Zoo~\cite{zooZooModel} have introduced early Text-to-CAD tools, reflecting the increasing demand for language-driven CAD. In parallel, a growing number of datasets have been released in recent years, enabling research on generative CAD with various input modalities and output formats~\cite{xu2024cad}. However,
existing approaches face inherent limitations due to their choice of representation. 

Most prior methods adopt task-specific command sequences~(see Appendix~\ref{supplementary:command_sequence_background}), which mimic the low-level construction history in CAD software. While this design aligns with how human users interact with CAD environments, it is not a natural output space for large language models. \revise{Command sequences are a custom token format that pretrained models have not seen, so a model must learn the output format entirely from task data, and prior methods such as Text2CAD train task-specific models from scratch. The sequences are also not human-readable, and they are tied to specific datasets and software conventions: a solid can only be obtained through dataset-specific conversion code.} On the other hand, boundary representation (B-rep) formats capture the geometry and topology of solid models in a mathematically rigorous way. Although B-rep is a standard in modern CAD systems, it operates at an extremely low level, encoding faces, edges, and vertices explicitly. This makes generation challenging: producing valid B-rep structures requires strict geometric consistency, and even minor numerical errors can lead to invalid or non-manifold models. Debugging or repairing such representations is often more complex than constructing the models in the first place.

These limitations highlight the need for an alternative representation that is both high-level and executable, while remaining compatible with the strengths of large pretrained models. To fill this gap, we ask a simple question: Why generate intermediate command sequences at all? Why not generate executable code that can directly produce a 3D model? 

Among the available options (see Appendix~\ref{supplementary: parametric_cad_example}), we choose CadQuery~\cite{cadquery_2024} as the target representation. Unlike FreeCAD~\cite{riegel2016freecad}, which is tightly coupled with a GUI-based environment and difficult to scale for automated generation, or OpenSCAD~\cite{machado2019parametric}, which relies on a domain-specific language with limited expressiveness, or PythonOCC, which exposes low-level OpenCascade primitives through verbose APIs, CadQuery strikes a favorable balance between usability and flexibility. As a pure Python package, it can be executed directly without external dependencies, making it lightweight and easy to integrate with modern ML pipelines. Beyond its practicality, CadQuery also offers unique advantages for research. Its high-level API provides direct support for fundamental modeling operations such as box, arc, circle, and extrude, as well as Boolean operations like union, cut, and intersection. These primitives can be flexibly combined to construct arbitrarily complex 3D models, while maintaining concise and readable code. This blend of readability, executability, and composability makes CadQuery particularly well suited as an output space for generative modeling. From an evaluation standpoint, it further enables direct traceability between code and geometry, allowing systematic debugging, automated execution that command sequences or B-rep representations cannot provide as naturally.

The use of CadQuery also offers two key advantages. First, modern large models are already good at generating Python code, and in many cases, they are capable of writing CadQuery directly. Combined with their emerging capabilities in multimodal understanding and spatial reasoning, these models are naturally well suited for mapping high-level inputs directly into 3D generative code. \revise{Because CadQuery is ordinary Python, it allows us} to directly build on pretrained models and adapt them efficiently through fine-tuning\revise{, without learning a new output format from task data alone}. Second, model scaling laws suggest that larger pretrained models inherently perform better~\cite{kaplan2020scaling}. This trend carries over to our task: \revise{within one model family trained under identical settings,} as model capacity increases, we observe more accurate CadQuery generation. In other words, improvements in foundation models transfer to \revise{Text-to-CadQuery generation, so that CadQuery generation} \revise{benefits from advances} in model size and capability.

Building on these insights, the central goal of this work is to demonstrate that CadQuery serves as \revise{an effective} output representation for generative CAD. In particular, we show that CadQuery \revise{matches task-specific command sequences in shape accuracy while producing readable programs that execute directly, without dataset-specific conversion code, and that its generation quality improves with model size}: as pretrained models grow larger and more capable, their ability to generate valid and accurate CadQuery improves correspondingly.

To pursue this vision, we take two key steps. First, we build on the Text2CAD~\cite{khan2024text2cad} dataset, which augments DeepCAD~\cite{wu2021deepcad} with natural language descriptions of 3D models. We further extend this dataset by annotating each sample with corresponding CadQuery code\revise{, obtained by deterministically translating its source CAD model and verified geometrically against it}. This results in a new dataset of approximately 170,000 text–CadQuery pairs, enabling direct training for Text-to-CadQuery generation. Second, we fine-tune \revise{four models of one family, Qwen2.5-0.5B, 1.5B, 3B and 7B~\cite{yang2024qwen2}, under identical settings, and we also train the 0.5B model on command sequences to isolate the effect of the output representation}. Our experiments reveal two important findings. Unlike prior work that relies on custom command sequences and requires training a task-specific decoder—even when paired with a pretrained BERT encoder, resulting in a total model size of around 363M parameters—our smallest model surpasses Text2CAD in CAD generation accuracy after \revise{about three hours} of training, compared to two full days required \revise{to train Text2CAD from scratch}. \revise{With the same backbone, CadQuery matches command sequences in shape accuracy and is slightly more accurate in absolute dimensions}\revise{; on an external benchmark with differently styled prompts, it also transfers better, which we regard as supporting evidence that it draws on pretrained knowledge}. Moreover, we observe a clear scaling trend: larger models yield higher generation accuracy. These results support our two core claims: (1) generating CadQuery directly \revise{matches task-specific command sequences in accuracy on the same pretrained backbone, while producing readable, editable programs that execute directly, without dataset-specific conversion code}; and (2) larger pretrained models do indeed bring stronger CAD generation ability. Building on these findings, we summarize our main contributions as follows:
\begin{itemize}
    \item We formulate text-to-CAD as direct generation of executable CadQuery code, which yields human-readable programs that run without dataset-specific conversion code. \revise{Through a controlled comparison with command sequences on the same pretrained backbone, we show that this representation matches them in shape accuracy and is more accurate in absolute dimensions, with \rtwo{preliminary} evidence that it transfers better to an external benchmark.}
 
    \item We construct a dataset of 170,000 text–CadQuery pairs by extending the Text2CAD dataset with executable CadQuery annotations aligned to each prompt\revise{, derived deterministically from the source CAD models and validated geometrically}.  This enables direct training for Text-to-CadQuery generation and supports future research in code-based 3D modeling.
    
    \item We perform a comprehensive evaluation across \revise{four sizes of one model family, Qwen2.5 from 0.5B to 7B, trained under identical settings}. Our results reveal a consistent scaling trend: larger models produce more accurate and executable CadQuery code, confirming that increased model capacity improves symbolic 3D generation.
\end{itemize}

The rest of this paper is organized as follows. Section~\ref{sec:related_work} reviews related work. Section~\ref{sec:method} details our method, including dataset annotation process, pretrained model selection, and fine-tuning setup. Section~\ref{sec:results} presents evaluation metrics and experimental results. Section~\ref{sec:discussion} discusses implications, limitations and future directions. Finally, Section~\ref{sec:conclusion} concludes the paper.

\section{Related Work}\label{sec:related_work}
\subsection{Datasets and Methods for Generative CAD}
Early CAD datasets mainly consisted of static 3D models, often in boundary representation (B-rep) format, without associated modeling steps or input–output pairs. While such datasets, such as ABC~\cite{koch2019abc}, enabled progress in geometric deep learning, they were not suitable for generative modeling tasks that require capturing design intent or procedural histories. To bridge this gap, more recent datasets have introduced labeled modeling sequences, natural language descriptions, images, or sketches, making it possible to train conditional generative models. These datasets can be broadly categorized by input type: image-based (renderings, sketches), text-based (instructions or descriptions), point clouds, or multimodal combinations. Output formats also vary considerably, ranging from command sequences and B-rep structures to script-based CAD representations. This diversity of modalities has made CAD an increasingly popular benchmark for testing generative modeling methods. A comprehensive overview is provided in Tab.~\ref{tab:review}.

Within this spectrum, command sequence based datasets such as SketchGraphs~\cite{seff2020sketchgraphs} and DeepCAD~\cite{wu2021deepcad} have played a central role. These datasets encode CAD operations (e.g., line, arc, extrusion) into sequential commands that mirror the construction history in CAD software\revise{, and Drawing2CAD~\cite{qin2025drawing2cad} applies the same representation to vector drawings}. While intuitive from a software perspective, this representation is not a natural output space for large language models. \revise{Command sequences form a custom token format that pretrained models have not seen, and because they are tightly coupled with specific CAD platforms and datasets, methods built on them are typically trained from scratch for each dataset.} B-rep–based datasets such as ABC~\cite{koch2019abc}, SolidGen~\cite{jayaraman2022solidgen}, and BrepGen~\cite{xu2024brepgen} adopt a mathematically rigorous representation of solids, encoding faces, edges, and topological relations\revise{, and recent methods such as BRep-GD~\cite{qin2026brep} and HiFi-BRep~\cite{hou2026hifi} improve the robustness and fidelity of B-rep generation}. Although B-rep is the industrial standard in CAD kernels, it is extremely low-level and difficult to generate. Ensuring manifoldness and geometric consistency during learning is non-trivial, and even minor numerical errors may yield invalid or incomplete models. As a result, B-rep generation remains computationally expensive and error-prone, limiting its practical utility for scalable generative modeling. 

A third line of work explores script-based CAD representations, including Query2CAD~\cite{badagabettu2024query2cad} and LLM4CAD~\cite{sun2025large}. These methods leverage executable CAD scripts written in languages such as FreeCAD macros or CadQuery code. Script-based approaches combine interpretability with executability: models directly produce code that can be executed to yield 3D geometry, facilitating debugging and evaluation. However, existing datasets in this space are still limited. Query2CAD, for instance, contains only 57 examples, making it more of a proof-of-concept than a benchmark. LLM4CAD annotates approximately 5,000 CadQuery samples, but the scope is restricted to five families of mechanical parts, limiting generalization. These limitations underscore the need for a larger, more diverse dataset that leverages the advantages of code-based CAD representations—precisely the gap that motivates our construction of a 170,000-sample text–CadQuery corpus.

Although multimodal inputs, such as combining natural language with images or sketches, are often assumed to improve performance, recent studies suggest that this is not always the case. In particular, LLM4CAD~\cite{li2025llm4cad} reports that, in the current stage where datasets remain relatively small and models are not yet highly specialized, using text alone as the input achieves the highest accuracy. It even outperforms both image-based pipelines and multimodal combinations. This finding is consistent with the broader observation that large language models are most mature when operating in the text domain, making natural language the most reliable and scalable entry point for generative CAD research. Building on this insight, we adopt text as the sole input modality and focus our contributions on the output representation. Among existing text-to-CAD methods, Text2CAD~\cite{khan2024text2cad} is the first work to extend DeepCAD~\cite{wu2021deepcad} with textual annotations, pairing natural language descriptions with command-sequence outputs. Its design provides a natural point of comparison for our work: we use exactly the same text inputs as Text2CAD, but replace its task-specific command-sequence outputs with executable CadQuery code. \revise{Because Text2CAD also differs from our models in architecture and pretraining, we additionally train the same pretrained model on its command sequences to isolate the effect of using CadQuery as the output format.} As a result, Text2CAD serves as the primary baseline throughout our experiments.

\begin{table}[!t]
\vspace{-1mm}
\centering
\caption{Overview of CAD Datasets by Input Condition, CAD Representation, and \revise{Dataset Size}}
\label{tab:review}
\renewcommand{\arraystretch}{1.1}
\setlength{\tabcolsep}{2.5pt}
\footnotesize
\begin{tabular}{@{}llcll@{}}
\toprule
\textbf{Input} & \textbf{Dataset} & \textbf{Year} & \textbf{Output} & \textbf{Size} \\
\midrule
\multirow{9}{*}{Uncond.}
 & ABC~\cite{koch2019abc} & 2019 & B-rep & 1M+ \\
 & SketchGraphs~\cite{seff2020sketchgraphs} & 2020 & Command seq. & 15M+ \\
 & DeepCAD~\cite{wu2021deepcad} & 2021 & Command seq. & 179k \\
 & Fusion360~\cite{willis2021fusion} & 2021 & Command seq. & 8.6k \\
 & CC3D-Ops~\cite{dupont2022cadops} & 2022 & B-rep & 37k+ \\
 & SkexGen~\cite{xu2022skexgen} & 2022 & Command seq. & $\sim$100k \\
 & CADParser~\cite{zhou2023cadparser} & 2023 & Command seq. & 40k+ \\
 & SolidGen~\cite{jayaraman2022solidgen} & 2023 & B-rep & $\sim$100k \\
 & BrepGen~\cite{xu2024brepgen} & 2024 & B-rep & $\sim$100k \\
\midrule
\multirow{3}{*}{Image}
 & OpenECAD~\cite{yuan2024openecad} & 2024 & Command seq. & 200k+ \\
 & Img2CAD (GPT4-V)~\cite{you2025img2cad} & 2025 & Command seq. & 4.6k \\
 & Img2CAD (SVG)~\cite{chen2025img2cad} & 2025 & Command seq. & 209k \\
\midrule
\multirow{2}{*}{Text}
 & Query2CAD~\cite{badagabettu2024query2cad} & 2024 & Python macro & 57 \\
 & Text2CAD~\cite{khan2024text2cad} & 2024 & Command seq. & 158k+ \\
\midrule
\multirow{3}{*}{Multi-modal}
 & Objaverse-XL~\cite{deitke2023objaverse} & 2023 & Mesh & 10M+ \\
 & OmniCAD~\cite{xu2024cad} & 2024 & Command seq. & 453k \\
 & LLM4CAD~\cite{sun2025large} & 2025 & CadQuery & $\sim$5k \\
\midrule
Sketch & Free2CAD~\cite{li2022free2cad} & 2022 & Command seq. & 210k+ \\
\midrule
\multirow{2}{*}{Point cloud}
 & Point2CAD~\cite{liu2024point2cad} & 2023 & B-rep & ABC subset \\
 & Point2Skh~\cite{wang2025point2skh} & 2024 & Sketch+extrude & Synthetic \\
\bottomrule
\end{tabular}
\end{table}

\subsection{General Capabilities of Large Models Enabling CAD Modeling} 

Parallel to dataset development, advances in foundation models have dramatically expanded the feasibility of generative CAD. Modern LLMs have acquired a broad set of general-purpose capabilities that can be directly leveraged for CAD tasks. State-of-the-art models such as GPT-4~\cite{achiam2023gpt} and Gemini~\cite{team2023gemini} are inherently multimodal, capable of integrating language, vision, and spatial information for cross-modal reasoning. Their emergent spatial reasoning abilities make them well suited to interpret geometric descriptions and maintain 3D coherence across long sequences of operations.

Equally important, LLMs good at structured code generation. Systems such as Qwen coder~\cite{hui2024qwen2} have shown that pretrained models can translate natural language into executable code across diverse domains. By analogy, parametric CAD models, often expressed as text-based command sequences or scripts, can be viewed as a specialized form of structured program synthesis. This connection suggests that pretrained language models, especially those with strong code generation priors, are inherently compatible with CAD script generation.

Building on this observation, several works have begun to test whether code-based CAD representations can directly benefit from large models. Query2CAD~\cite{badagabettu2024query2cad} demonstrates that GPT-4 can produce valid FreeCAD macros from textual descriptions, but the dataset size is too small for robust training. LLM4CAD~\cite{sun2025large} shows that CadQuery can serve as an effective output representation, but its dataset covers only a narrow set of mechanical parts. These studies highlight the promise of combining large pretrained models with code-based CAD, while also revealing the need for larger and more diverse datasets to validate this direction.

To fill these technical gaps, we extend CadQuery to a much larger and more diverse dataset and conduct a comprehensive evaluation across multiple pretrained models. Unlike command sequences or B-rep, CadQuery offers a high-level, executable, and LLMs friendly format that aligns with the strengths of pretrained language models. To \revise{test} this choice, we \revise{compare CadQuery with command sequences on the same pretrained backbone, and we} systematically evaluate text-to-CadQuery generation across models of varying scales, testing whether larger pretrained models can better leverage their code-generation and spatial reasoning abilities in this setting. Our results ultimately demonstrate that CadQuery is not only \revise{an effective} representation but also one \revise{whose generation quality improves} with advances in foundation models: As models grow in size and capability, their CadQuery generation improves accordingly. In this way, our work provides \revise{controlled evidence of a consistent scaling trend in} symbolic CAD generation, showing that Text-to-CadQuery generation can continuously benefit from progress in pretrained models. This positions CadQuery as a promising representation for future research in CAD generation, laying the foundation for scalable, generalizable, and multimodal AI-driven design.

\textbf{Concurrent work.} Independently and concurrently with our study, two efforts have also explored CadQuery as a generation target. CAD-Coder~\cite{doris2026cad} trains an open-source vision-language model that produces CadQuery code from images of CAD models, whereas a concurrent system of the same name~\cite{guan2026cad} generates CadQuery from text using chain-of-thought reasoning and reinforcement learning with a geometry-based reward. Together with ours, these works reflect a growing consensus that executable, Python-native CAD scripting is a natural output space for large pretrained models. Our contribution is complementary and distinct: rather than introducing a specialized training objective or a different input modality, we provide a large-scale text–CadQuery dataset\revise{, a controlled comparison with command sequences on the same backbone,} and a systematic evaluation of how generation quality scales across \revise{four sizes of one model family, Qwen2.5 from 0.5B to 7B}\revise{, on which} subsequent CAD-aware and reinforcement-learning-based methods can build.

\section{Method}\label{sec:method}
\subsection{Data Annotation}\label{sec:annotation}
\textbf{Base Dataset Description}: We build our dataset on top of two foundational CAD datasets: DeepCAD~\cite{wu2021deepcad} and Text2CAD~\cite{khan2024text2cad}. DeepCAD contains 178,238 CAD models represented as command sequences, each describing a sketch-and-extrude construction process used in real-world CAD software. The dataset covers a wide range of user-generated mechanical parts from the Onshape platform, and supports parametric commands such as line, arc, circle, and various extrusion types. Text2CAD extends DeepCAD by annotating each model with multiple levels of natural language descriptions, ranging from abstract shape summaries to detailed parametric instructions. The resulting dataset includes approximately 170,000 models and over 600,000 prompts, which reflect diverse prompt styles and levels of detail.

\begin{figure*}[h]
  \centering
  \includegraphics[width=0.98\linewidth]{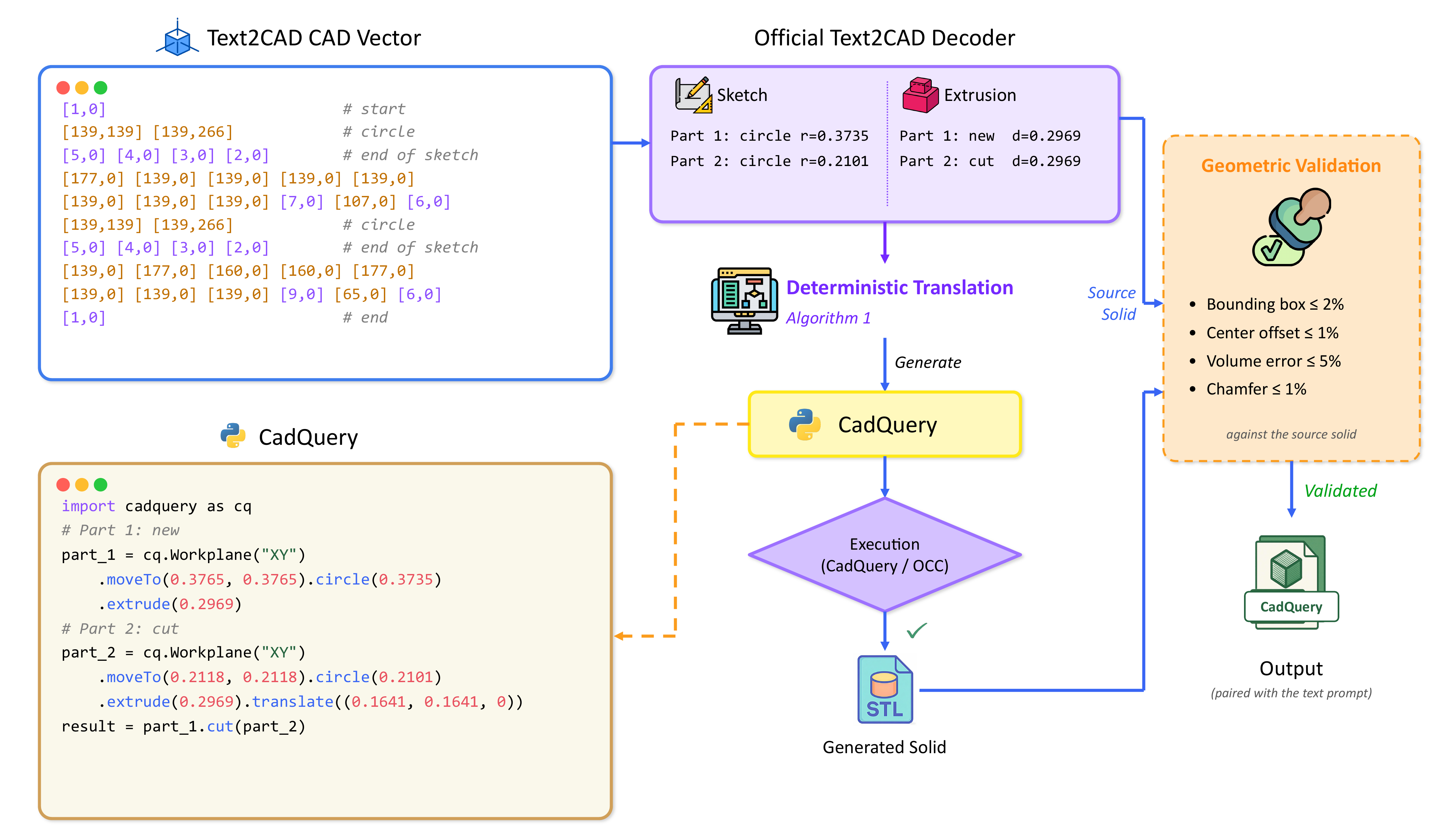}
  \vspace{-2mm}
  \caption{Overview of the \revise{deterministic} data annotation pipeline.}
  \label{fig:flowchart}
\end{figure*}

\begingroup
\textbf{Deterministic Annotation}: Instead of asking a language model to write CadQuery code, we derive each CadQuery program directly from the source CAD model (Fig.~\ref{fig:flowchart}). For every model, Text2CAD~\cite{khan2024text2cad} stores its construction history as a quantized CAD vector (see Appendix~\ref{supplementary:command_sequence_background}), which is the output that the Text2CAD Transformer is trained to generate and which the official Text2CAD decoder converts into a solid model. \revise{This decoder is the dataset-specific conversion code released with Text2CAD that turns a command sequence into a 3D model.} This vector fully specifies the model as an ordered list of sketch-and-extrude steps. Each sketch consists of faces bounded by closed loops of lines, arcs and circles, and each extrusion is defined by a sketch plane (origin and orientation), one or two extents along the plane normal, and a Boolean operation. Because this specification determines the CadQuery program completely, the translation can be carried out by a short deterministic program, and no language model is involved.

Algorithm~\ref{alg:cadvec2cq} summarizes the conversion. We first decode the CAD vector with the official Text2CAD decoder, so that the conversion operates on exactly the geometry that Text2CAD itself builds. We then follow the decoder's own construction logic: an extrusion that produces zero volume is skipped, and when a later extrusion cannot be built, the model is truncated at that step. For each remaining step, every sketch face is drawn on the XY workplane, with its loops ordered by bounding-box area so that the outer boundary precedes the inner holes. Lines, arcs and circles become \texttt{lineTo}, \texttt{threePointArc} and \texttt{circle} calls, respectively. A two-sided extrusion is written as a single extrusion over the interval it covers along the normal, by shifting the start of the profile to the lower end of the interval. The extruded part is then placed in the global frame by three rotations about the X, Y and Z axes, which reproduce the orientation of the sketch plane, followed by a translation to its origin. Finally, the parts are combined in order: \textit{new} and \textit{join} become a union, \textit{cut} a subtraction, and \textit{intersect} is written as $A-(A-B)$. Coordinates are printed with four decimals, or six when two distinct coordinates of the model lie closer than $5\times10^{-4}$, so that rounding never merges distinct vertices. A complete example is given in Appendix~\ref{supplementary: dataset_examples}.

\begin{algorithm}[!t]
\caption{CAD vector to CadQuery conversion}
\label{alg:cadvec2cq}
\begin{algorithmic}[1]
\REQUIRE CAD vector $v$ of one Text2CAD sample
\ENSURE CadQuery program $P$
\STATE $\{(S_i, E_i)\}_{i=1}^{N} \leftarrow \textsc{OfficialDecode}(v)$ \COMMENT{sketches and extrusions}
\STATE $\mathcal{U} \leftarrow \emptyset$
\FOR{$i = 1$ \TO $N$}
  \IF{$\textsc{Build}(S_i, E_i)$ fails}
    \STATE \textbf{if} $i = 1$ \textbf{then} reject sample \textbf{else} \textbf{break}
  \ENDIF
  \STATE \textbf{if} the built body has non-zero volume \textbf{then} $\mathcal{U} \leftarrow \mathcal{U} \cup \{i\}$
\ENDFOR
\STATE $d \leftarrow 4$ if the minimum coordinate spacing is $\ge 5\times10^{-4}$, else $6$
\FOR{each $i \in \mathcal{U}$}
  \STATE $[l, h] \leftarrow$ interval along the normal covered by the two extents of $E_i$
  \FOR{each face $F$ of $S_i$}
    \STATE emit \texttt{Workplane("XY")}; for each loop of $F$, in decreasing bounding-box area:
    \STATE \quad circle $\rightarrow$ \texttt{moveTo(c).circle(r)}
    \STATE \quad else $\rightarrow$ \texttt{moveTo}; per curve \texttt{lineTo} or \texttt{threePointArc}; \texttt{close()}
    \STATE emit \texttt{extrude($h-l$)}; union with the previous faces of step $i$
  \ENDFOR
  \STATE $(\alpha,\beta,\gamma) \leftarrow$ extrinsic XYZ Euler angles of the sketch-plane frame
  \STATE emit \texttt{rotate} by $\alpha,\beta,\gamma$ about X, Y, Z, then \texttt{translate} to origin $+\, l\cdot$normal
\ENDFOR
\STATE combine parts in order: new/join $\rightarrow$ \texttt{union}; cut $\rightarrow$ \texttt{cut}; intersect $\rightarrow$ \texttt{cut(cut(\,))}
\STATE all numbers are printed with $d$ decimals
\RETURN $P$
\end{algorithmic}
\end{algorithm}

This construction has three consequences. First, each program is equivalent to its source model by construction. Second, the CadQuery targets, the command-sequence targets of Text2CAD, and the reference solids used for evaluation are all derived from the same CAD vector, so the two output representations describe identical geometry. This is what allows the controlled representation comparison in Sec.~\ref{sec:results}. Because the CAD vector is quantized, a value in the program may differ slightly from the value stated in the prompt (for example, 0.2969 for a stated height of 0.3) so that the program always matches the Text2CAD target exactly. Third, no language model takes part in annotation, so dataset construction is independent of the vision-language evaluation in Sec.~\ref{sec:results}.

\begin{table}[!b]
\vspace{-1mm}
\centering
\caption{Geometry-level validation of CadQuery annotations}
\label{tab:annotation_audit}
\renewcommand{\arraystretch}{1.15}
\begin{tabular}{@{}lr@{}}
\toprule
\textbf{Check} & \textbf{Pass rate} \\
\midrule
\multicolumn{2}{@{}l}{\textit{All annotations}} \\
Executes to a valid solid                   & 99.64\% \\
Bounding-box max-axis error $\le 2\%$        & 98.82\% \\
Center offset $\le 1\%$ of diagonal          & 99.48\% \\
Volume error $\le 5\%$                       & 99.37\% \\
\textbf{All three geometric checks}          & \textbf{98.71\%} \\
Equal solid count                            & 91.90\% \\
Median bounding-box / volume error           & 0.0002 / 0.0003 \\
\midrule
\multicolumn{2}{@{}l}{\textit{Stratified audit sample} ($n=1{,}000$)} \\
Chamfer distance $\le 1\%$ of diagonal       & 96.5\% \\
\textbf{Three geometric checks $+$ Chamfer}  & \textbf{95.0\%} \\
Median solid IoU                             & 1.00 \\
\bottomrule
\end{tabular}
\end{table}

\textbf{Geometry-Level Validation}: We validate every annotation by executing it and comparing the resulting solid with the solid built by the official Text2CAD decoder from the same CAD vector (Table~\ref{tab:annotation_audit}). The two solids are compared in their shared source coordinate frame, without re-centering or re-scaling, so that errors in dimensions, position and scale are all detected. The acceptance thresholds are fixed in advance: a maximum per-axis bounding-box error of at most 2\%, a center offset of at most 1\% of the reference bounding-box diagonal, and a volume error of at most 5\%. Every model with a valid source solid remains in the denominator, and an annotation that fails to execute counts as a failure. Under these rules, 99.64\% of the annotations execute and 98.71\% pass all three geometric checks, with median bounding-box and volume errors of 0.0002 and 0.0003. As a topology check, the number of solids agrees for 91.90\% of the models. Most of the mismatches (92\%) still pass every geometric check: they arise because touching bodies are counted as a single solid by one kernel and as several by the other. Annotations that fail any of these checks are removed, and all experiments use the remaining data.

Mesh-based metrics are costlier, so we compute them on a frozen stratified sample of 1,000 models (800/100/100 from the training, validation and test splits), stratified by the number of extrusions, the profile type (single face, multiple faces, inner loops) and the use of cut or intersect operations. The Chamfer distance between 10,000 surface points sampled from each solid, normalized by the reference diagonal, is at most 1\% for 96.5\% of the sample. 95.0\% pass all four checks, and the median volumetric IoU is 1.00. The remaining cases are mostly measurement artifacts, namely tessellation differences between the two kernels and non-watertight meshes for which IoU is undefined, rather than translation errors.

\textbf{Relation to CAD-Coder}: The concurrent CAD-Coder~\cite{doris2026cad} also translates DeepCAD sequences into CadQuery. Our conversion was developed independently: it starts from the Text2CAD CAD vector and its official decoder, so that our labels, the Text2CAD targets and our evaluation reference share a single source, and it is verified geometrically against the source solids rather than by execution alone. The construction also differs: instead of sketching on a workplane positioned in space and splitting a two-sided extrusion into two, we draw each profile in its local frame before rotating and translating it into place, and write a two-sided extrusion as a single extrusion over the interval it covers.
\endgroup

\subsection{Pretrained Model Selection}
Our goal is not only to validate that CadQuery is \revise{an effective output format compared with} task-specific command sequences, but also to investigate whether larger pretrained models are more capable of producing accurate, executable CadQuery code. In selecting models for fine-tuning, we consider both parameter scale and prior familiarity with CadQuery.

First, we assess whether the pretrained model demonstrates familiarity with CadQuery, as reflected in its ability to generate simple scripts without task-specific tuning. To assess this, we surveyed several popular open-source models including DeepSeek, Qwen, Gemma, Llama, and Mistral. As a sanity check, we prompted each model with two simple queries: “Do you know what CadQuery is?” and “Give me a simple CadQuery example.” Most models returned reasonable explanations and generated code snippets that followed the correct CadQuery API structure, even though the outputs were not directly executable. This suggests that these models have already acquired certain understanding of the syntax and structure of CadQuery during pre-training. Surprisingly, the highly regarded Llama 3 series consistently failed both questions, exhibiting strong hallucinations and generating irrelevant or incorrect responses. Second, we take into account the scale of model parameters. Text2CAD Transformer uses a 340M BERT-Large~\cite{devlin2019bert} encoder and a 23M decoder trained from scratch, resulting in a total size of 363M. \revise{To compare model sizes without confounding them with differences in pretraining data, architecture or tokenizer, we use a single model family, Qwen2.5-Instruct~\cite{yang2024qwen2}, at four sizes: 0.5B, 1.5B, 3B and 7B. The 0.5B model is the closest in size to the Text2CAD Transformer. To isolate the effect of the output format, we also fine-tune the 0.5B model on command sequences under identical settings: the target is the Text2CAD CAD vector itself, serialized losslessly as text (its integer token pairs separated by semicolons), and generated sequences are decoded into solids by the official Text2CAD decoder. Both models therefore receive the same prompts, data split and reference solids.}

\subsection{Fine-Tuning Setup}
We split the full 170k-sample dataset into 90\% for training, 5\% for validation, and 5\% for testing. \revise{The split is made at the level of CAD models, so each model appears in exactly one split. The effect of geometrically identical models across splits is analyzed in Sec.~\ref{sec:results}.} Despite their varying parameter sizes, all models converge efficiently. For reference, Text2CAD Transformer trains a 23M decoder from scratch over two days on a single A100 GPU. In contrast, our smallest model (\revise{Qwen2.5-0.5B}) fine-tunes in \revise{about three hours}, and even our largest model (\revise{Qwen2.5-7B}) completes training within \revise{16} hours (see Tab.~\ref{tab:training_setup}). \revise{All models use the same protocol: full-parameter supervised fine-tuning (SFT) in BF16 with AdamW, a learning rate of $5\times10^{-5}$ with linear decay, two epochs, and an effective batch size of 12. When a model does not fit a micro-batch of four sequences, we accumulate gradients over 12 micro-batches of one. To assess the stability of training, the 0.5B CadQuery model is trained with three random seeds (42, 123 and 2026) and all other models use seed 42.}

\begingroup
\begin{table}[!t]
\vspace{-1mm}
\renewcommand{\arraystretch}{1.2}
\caption{Fine-tuning configurations for all models}
\label{tab:training_setup}
\centering
\begin{tabular}{cccc}
\toprule
\textbf{Model} & \textbf{Output} & \textbf{Batch Size} & \textbf{Training Time} \\
\midrule
Qwen2.5-0.5B & CadQuery     & 12 (4$\times$3)  & 3h 13m \\
Qwen2.5-0.5B & Command seq. & 12 (4$\times$3)  & 2h 53m \\
Qwen2.5-1.5B & CadQuery     & 12 (4$\times$3)  & 5h 10m \\
Qwen2.5-3B   & CadQuery     & 12 (1$\times$12) & 13h 52m \\
Qwen2.5-7B   & CadQuery     & 12 (1$\times$12) & 15h 28m \\
\midrule
\multicolumn{4}{c}{All: 1$\times$A100 80GB, full SFT, 2 epochs, learning rate 5e-5} \\
\bottomrule
\end{tabular}
\end{table}
\endgroup

\textbf{Tokenization and Context Length:} We tokenize each input-output pair into a single prompt with instruction–response formatting. Across the 170,000 samples, token statistics \revise{with the Qwen2.5 tokenizer} show a minimum length of \revise{168}, a maximum of \revise{2,343}, a mean of \revise{428.1}, and a median of \revise{331} tokens. Notably, \revise{95.96}\% of all samples fall below 1024 tokens. \revise{We set the context length to 4,096 for all models, so that no sample is truncated.} All textual queries in the dataset explicitly specify numeric parameters (e.g., dimensions, radii, and lengths), enabling precise parametric CAD generation rather than purely high-level shape descriptions. Representative text--CadQuery pairs are provided in Appendix~\ref{supplementary: dataset_examples}.

Since our task focuses on generating accurate 3D models, the primary concern is whether the generated code produces the correct geometry upon execution, rather than its textual form. However, integrating geometric accuracy into the loss function is non-trivial and computationally expensive. Therefore, during training, we adopt the standard token-level cross-entropy loss. \revise{At test time, each model produces a single output per prompt by greedy decoding.} Detailed evaluation of 3D shape correctness, including geometry-aware metrics, is performed post-training, as described in Sec.~\ref{sec:results}.

\begingroup
\section{Results and Evaluation}\label{sec:results}

\subsection{Evaluation Protocol}\label{sec:protocol}
\textbf{Test Set and Inputs:} The test split contains 8,744 CAD models. Each model is described by its intermediate-level Text2CAD prompt, and every system receives exactly the same text. The reference solid is always the one built by the official Text2CAD decoder from the source CAD vector. This is the geometry that the CadQuery annotations reproduce (Sec.~\ref{sec:annotation}), that the command-sequence targets encode, and that Text2CAD is trained to generate, so all systems are scored against the same target.

\textbf{Invalid Outputs:} An output is invalid if it cannot be parsed, executed or decoded, or if it does not yield a valid solid. CadQuery programs are executed in a sandbox with a fixed time limit and command sequences are decoded by the official Text2CAD decoder. Invalid outputs remain in the denominator of every metric: they receive F1 $=$ IoU $=0$, count as failures for every pass rate, and count as ``No'' in the VLM evaluation. Chamfer distance is undefined for an invalid output, so we compute it only on the test items for which every compared model produces a valid solid.

\textbf{Metrics:} We group the metrics into four families, matching the column groups of Table~\ref{tab:main_results}.

\textit{Validity.} The Invalid Rate (IR) is the percentage of test items whose output is invalid.

\textit{Shape similarity (normalized).} Following prior work, Chamfer Distance (CD)~\cite{fan2017point}, F1 and volumetric IoU~\cite{wu20153d} compare shapes after normalizing each mesh independently, so they measure shape similarity but ignore absolute size and position. For CD and F1, each mesh is centered at its centroid and scaled so that its longest bounding-box side is 1, and 10,000 points are sampled uniformly from its surface. CD is the symmetric mean squared nearest-neighbor distance,
\begin{equation}
\mathrm{CD}(P, Q) = \frac{1}{|P|} \sum_{p \in P} \min_{q \in Q} \|p - q\|^2 + \frac{1}{|Q|} \sum_{q \in Q} \min_{p \in P} \|q - p\|^2
\label{eq:chamfer}
\end{equation}
where $P$ and $Q$ are the point sets sampled from the predicted and reference shapes. We report it $\times 10^3$. F1 uses the same points: a point counts as correct if its nearest neighbor in the other set lies within $\tau = 0.02$, and
\begin{equation}
    \mathrm{F1} = \frac{2 \cdot \text{Precision} \cdot \text{Recall}}{\text{Precision} + \text{Recall}}.
\end{equation}
For IoU, both meshes are normalized to the unit cube $[0,1]^3$ and voxelized with a pitch of 0.02; with $V_1$ and $V_2$ the occupied voxels,
\begin{equation}
    \mathrm{IoU} = \frac{|V_1 \cap V_2|}{|V_1 \cup V_2|}.
\end{equation}

\textit{Engineering accuracy (raw coordinates).} Two shapes can be similar after normalization and still describe different parts. We therefore also compare the prediction and the reference in their original coordinates, without any centering or scaling, using thresholds fixed before evaluation. \textit{Dim} requires each of the three axis-aligned bounding-box dimensions to be within 5\% of the reference. \textit{Vol} requires the volume to be within 10\% of the reference. \textit{Center} requires the bounding-box center to lie within 5\% of the reference diagonal. \textit{Topology} requires the same number of connected bodies, the same Euler characteristic and the same genus. \textit{Strict} requires all four checks and, in addition, a valid B-rep in the CAD kernel. Each metric is the percentage of test items that pass.

\textit{VLM judgment.} We render the reference and the prediction with Blender~\cite{blender} under the same camera and lighting and place them side by side. Following the use of vision-language models as evaluators of 3D generation~\cite{wu2024gpt}, a vision-language model, Gemini 3 Flash (\texttt{gemini-3-flash-preview}), receives the image together with the text description and answers whether the prediction plausibly represents the same design (Yes/No). We report the percentage of Yes answers. Because this judgment is lenient about dimensional deviations (Sec.~\ref{sec:evaluators}), we use it only as an auxiliary perceptual metric; the deterministic metrics above are the primary evidence. The model is not used anywhere in data annotation.

\textbf{Statistics:} All comparisons between two models use a paired bootstrap over the same test items (10,000 resamples). A difference is called significant when its 95\% confidence interval (CI) excludes zero. To measure training variance, the 0.5B CadQuery model is trained with three seeds.

\textbf{Duplicates:} DeepCAD contains many repeated simple parts. We flag a test item as a duplicate when a training item has an identical source program or an identical reference B-rep signature (bounding-box extents, volume, and numbers of solids, faces and edges). This applies to 43.7\% of the test items. We therefore report every result on both the full test set and the remaining de-duplicated test set of 4,923 items.

\subsection{Controlled Representation Comparison}\label{sec:repr}

\begin{table*}[!t]
\vspace{-1mm}
\centering
\caption{Comparison of output representations and model sizes}
\label{tab:main_results}
\renewcommand{\arraystretch}{1.2}
\setlength{\tabcolsep}{4pt}
\resizebox{\textwidth}{!}{%
\begin{tabular}{@{}lcccccccccc@{}}
\toprule
& \textbf{Validity} & \multicolumn{3}{c}{\textbf{Shape similarity (normalized)}} & \multicolumn{5}{c}{\textbf{Engineering accuracy (raw coordinates)}} & \textbf{VLM} \\
\cmidrule(lr){2-2}\cmidrule(lr){3-5}\cmidrule(lr){6-10}\cmidrule(l){11-11}
\textbf{Model} & \textbf{IR}\,\%$\downarrow$ & \textbf{F1} mean / med$\uparrow$ & \textbf{IoU} mean / med$\uparrow$ & \textbf{CD}$\times10^3$ mean / med$\downarrow$
& \textbf{Dim}$\le$5\%$\uparrow$ & \textbf{Vol}$\le$10\%$\uparrow$ & \textbf{Center}$\uparrow$ & \textbf{Topology}$\uparrow$ & \textbf{Strict}$\uparrow$ & \textbf{Gemini}$\uparrow$ \\
\midrule
\multicolumn{11}{@{}l}{\textit{Full test set} ($n=8{,}744$)} \\
Qwen2.5-0.5B, command seq. & 2.66 & 0.674 / 0.886 & 0.691 / 0.865 & 14.26 / 0.267 & 67.51 & 57.19 & 73.24 & 78.89 & 42.42 & 61.02 \\
Qwen2.5-0.5B, CadQuery     & 4.65 & 0.668 / 0.896 & 0.689 / 0.875 & 14.22 / 0.253 & 69.54 & 59.11 & 72.34 & 76.83 & 43.08 & 60.47 \\
Qwen2.5-1.5B, CadQuery     & 4.05 & 0.672 / 0.905 & 0.690 / 0.875 & 14.25 / 0.253 & 69.83 & 59.62 & 72.43 & 77.94 & 42.90 & 62.01 \\
Qwen2.5-3B, CadQuery       & 3.19 & 0.702 / 0.944 & 0.719 / 0.928 & 12.57 / 0.216 & 73.91 & 63.57 & 76.21 & 78.34 & 47.22 & 65.29 \\
Qwen2.5-7B, CadQuery       & \textbf{2.33} & \textbf{0.725} / \textbf{0.960} & \textbf{0.742} / \textbf{0.957} & \textbf{10.96} / \textbf{0.206}
                           & \textbf{77.06} & \textbf{65.96} & \textbf{78.92} & \textbf{79.40} & \textbf{50.06} & \textbf{67.97} \\
\midrule
\multicolumn{11}{@{}l}{\textit{De-duplicated test set} ($n=4{,}923$)} \\
Qwen2.5-0.5B, command seq. & 4.13 & 0.571 / 0.600 & 0.591 / 0.651 & 18.53 / 1.901 & 56.12 & 43.42 & 64.58 & 69.63 & 25.20 & 46.58 \\
Qwen2.5-0.5B, CadQuery     & 6.90 & 0.566 / 0.603 & 0.589 / 0.661 & 18.30 / 1.607 & 59.99 & 46.35 & 63.87 & 66.86 & 26.60 & 46.06 \\
Qwen2.5-1.5B, CadQuery     & 6.18 & 0.570 / 0.623 & 0.590 / 0.660 & 18.15 / 1.770 & 60.32 & 47.17 & 63.80 & 68.62 & 26.86 & 47.30 \\
Qwen2.5-3B, CadQuery       & 4.82 & 0.597 / 0.671 & 0.618 / 0.715 & 17.32 / 1.222 & 65.10 & 50.78 & 68.03 & 68.42 & 29.92 & 51.17 \\
Qwen2.5-7B, CadQuery       & \textbf{3.71} & \textbf{0.617} / \textbf{0.714} & \textbf{0.639} / \textbf{0.741} & \textbf{15.53} / \textbf{0.912}
                           & \textbf{67.97} & \textbf{51.99} & \textbf{70.64} & \textbf{69.82} & \textbf{31.51} & \textbf{53.52} \\
\bottomrule
\end{tabular}}
\end{table*}

The first two rows of each block in Table~\ref{tab:main_results} compare CadQuery and command sequences with the same backbone (Qwen2.5-0.5B), training protocol, data split and reference geometry, so that the output representation is the only difference. \revise{In the table, arrows indicate whether higher or lower values are better, and bold marks the best value in each column.}

On shape similarity, the two representations are indistinguishable: none of the differences in F1, IoU, CD or the Gemini judgment is significant (e.g., F1 $-0.006$, 95\% CI $[-0.013, +0.001]$). The raw-coordinate metrics reveal a trade-off. CadQuery matches the reference dimensions and volume more often (Dim $+2.02$ points, $[+0.95, +3.10]$. Vol $+1.91$ points) and among valid outputs its median dimensional error is 0.0013 against 0.0017. Command sequences, in turn, are invalid less often (IR $-1.99$ points, $[-2.63, -1.35]$) and match the reference topology more often ($+2.06$ points). The center and strict pass rates do not differ significantly. The lower validity of CadQuery is expected: a CadQuery program fails as a whole if any API call or kernel operation fails (e.g., an empty selection or an invalid Boolean operation), whereas a command sequence follows a short fixed grammar that the decoder interprets directly. The topology gap plausibly has a related cause, since small kernel-level deviations, such as faces that touch instead of merging, change the number of bodies or the genus without changing the shape.

These conclusions are not an artifact of a single training run. Across three seeds (42, 123 and 2026), the 0.5B CadQuery model obtains IR $4.35\pm0.31$\%, F1 $0.670\pm0.002$, IoU $0.690\pm0.002$, CD $14.11\pm0.04$, Dim $69.74\pm0.26$\%, Vol $59.07\pm0.19$\%, Center $72.46\pm0.18$\%, Topology $77.17\pm0.29$\%, Strict $43.13\pm0.17$\% and a Gemini Yes rate of $60.64\pm0.17$\% (mean $\pm$ standard deviation; CD on the items for which all models are valid), and every seed is above the command model on Dim and Vol and below it on validity. The same picture holds on the de-duplicated test set: absolute scores are lower for both representations, which confirms that duplicates inflate absolute performance, but shape similarity remains indistinguishable, CadQuery remains more accurate in raw coordinates, and command sequences remain more often valid.

In summary, when everything except the representation is held fixed, CadQuery reaches the same shape accuracy as task-specific command sequences and slightly better dimensional accuracy, at the cost of a \rtwo{markedly higher invalid rate: on the de-duplicated test set it rises from 4.13\% to 6.90\%, a relative increase of about 67\% (2.66\% to 4.65\% on the full test set). Keeping an entire program executable is clearly harder for the model with CadQuery than with command sequences, and this trade-off should be taken into account when choosing the representation.} The value of CadQuery therefore lies less in raw accuracy than in what it adds at no \revise{cost in shape accuracy}. First, its outputs are standard Python programs that engineers can read, edit and reuse. Second, they execute directly in an open-source CAD kernel and export to STEP without \revise{dataset-specific conversion code}. Third, \revise{unlike a custom command format,} they let the model draw on the code knowledge acquired during pretraining\revise{. In distribution, where abundant training data is available, this prior knowledge brings no measurable gain in shape accuracy. However, on an external benchmark with differently styled prompts, CadQuery transfers better than command sequences (Sec.~\ref{sec:external}), which we regard as supporting evidence of this effect.}

\subsection{Effect of Model Size}\label{sec:scaling}
\begin{figure*}[!t]
  \centering
  \includegraphics[width=0.98\linewidth]{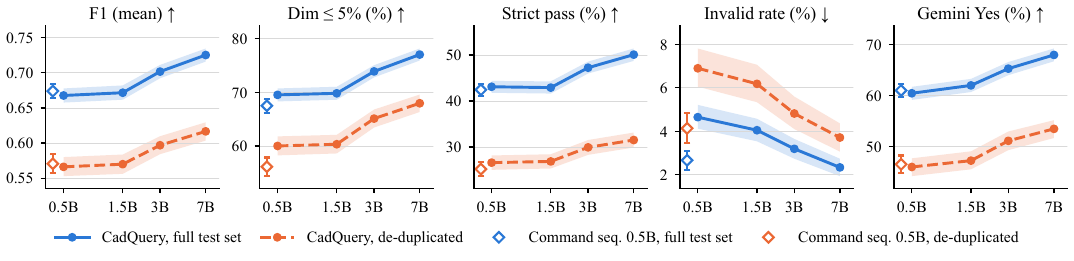}
  \vspace{-2mm}
  \caption{CadQuery generation quality across model sizes}
  \label{fig:scaling}
\end{figure*}

The CadQuery rows of Table~\ref{tab:main_results} cover four sizes of the same model family, trained under the identical protocol, and Fig.~\ref{fig:scaling} plots the main metrics against model size with 95\% confidence bands. Quality improves from 0.5B to 7B on every metric: IR halves from 4.65\% to 2.33\%, mean F1 rises from 0.668 to 0.725, Dim from 69.54\% to 77.06\%, Strict from 43.08\% to 50.06\%, and the Gemini judgment from 60.47\% to 67.97\%. The share of generated programs that are identical to the reference program grows from 25.8\% to 33.6\%.

The improvement is not uniform. From 0.5B to 1.5B most differences are small and not significant, whereas from 1.5B to 3B and from 3B to 7B almost every metric improves significantly. For example, F1 rises by $+0.030$ ($[+0.024, +0.037]$) and then by $+0.024$ ($[+0.018, +0.030]$). These gaps are more than ten times larger than the seed-to-seed standard deviation of the 0.5B model, so single-seed results for the larger models are unlikely to change this conclusion, assuming a comparable training variance at those sizes. The same trend holds on the de-duplicated test set.

\revise{Within one model family, with fixed data and training, larger models generate more accurate and more often valid CadQuery code, and the 7B model is the best on every metric. We study models up to 7B because this is the largest size that can be fully fine-tuned under our protocol on a single 80~GB GPU, and models of this size can also be deployed locally in design and manufacturing environments.} Fig.~\ref{fig:qualitative} shows selected examples on which the output improves as the model grows.

\begin{figure*}[!t]
  \centering
  \includegraphics[width=0.98\linewidth]{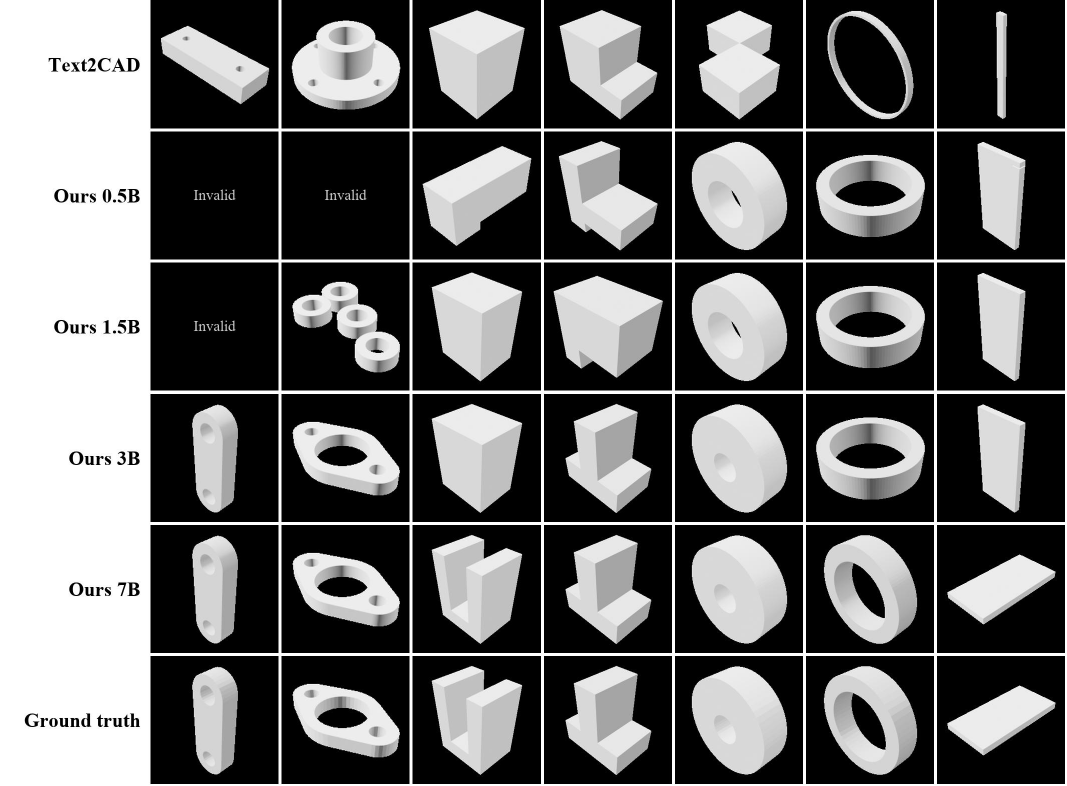}
  \vspace{-2mm}
  \caption{Selected qualitative examples on held-out test items shared with Text2CAD}
  \label{fig:qualitative}
\end{figure*}

\subsection{Comparison with Text2CAD and an External Benchmark}\label{sec:external}
\begin{table*}[!t]
\vspace{-1mm}
\centering
\caption{Comparison with Text2CAD and evaluation on CADTestBench}
\label{tab:external}
\renewcommand{\arraystretch}{1.2}
\setlength{\tabcolsep}{4pt}
\resizebox{\textwidth}{!}{%
\begin{tabular}{@{}lccccccc|cccc@{}}
\toprule
& \multicolumn{7}{c|}{\textbf{(a) Common held-out set} ($n=246$)} & \multicolumn{4}{c}{\textbf{(b) CADTestBench} ($n=82$)} \\
\cmidrule(lr){2-8}\cmidrule(l){9-12}
& & & & & & & & \multicolumn{2}{c}{Abstract prompt} & \multicolumn{2}{c}{Detailed prompt} \\
\textbf{Model} & \textbf{IR}\,\%$\downarrow$ & \textbf{F1} mean / med$\uparrow$ & \textbf{IoU} mean / med$\uparrow$ & \textbf{Dim}$\le$5\%$\uparrow$ & \textbf{Vol}$\le$10\%$\uparrow$ & \textbf{Strict}$\uparrow$ & \textbf{Gemini}$\uparrow$
& \textbf{IR}\,\%$\downarrow$ & \textbf{Pass}\,\%$\uparrow$ & \textbf{IR}\,\%$\downarrow$ & \textbf{Pass}\,\%$\uparrow$ \\
\midrule
Text2CAD (official, re-evaluated) & 3.25 & 0.336 / 0.170 & 0.383 / 0.300 & 22.0 & 22.4 & 11.4 & 28.0 & 3.66  & 59.23 & 1.22 & 44.03 \\
Qwen2.5-0.5B, command seq.      & 1.63 & 0.648 / 0.783 & 0.693 / 0.806 & 68.7 & 56.1 & 40.7 & 60.2 & 21.95 & 56.99 & 6.10 & 52.65 \\
Qwen2.5-0.5B, CadQuery          & 6.10 & 0.630 / 0.802 & 0.678 / 0.838 & 69.9 & 55.3 & 40.2 & 57.7 & 0.00  & 73.98 & 3.66 & 64.90 \\
Qwen2.5-1.5B, CadQuery          & 4.88 & 0.642 / 0.779 & 0.688 / 0.849 & 71.1 & 58.9 & 41.5 & 58.1 & 0.00  & 71.53 & 2.44 & 67.59 \\
Qwen2.5-3B, CadQuery            & 2.85 & 0.686 / 0.887 & 0.726 / 0.895 & 75.6 & 63.4 & 46.7 & 63.0 & 1.22  & 77.27 & 1.22 & 68.02 \\
Qwen2.5-7B, CadQuery            & 2.85 & \textbf{0.692} / \textbf{0.898} & \textbf{0.738} / \textbf{0.926} & \textbf{76.8} & \textbf{65.9} & \textbf{47.6} & \textbf{65.4}
                                & 0.00 & \textbf{79.07} & 1.22 & \textbf{68.39} \\
\midrule
\multicolumn{12}{@{}l}{\textbf{Paired differences} (95\% CI)} \\
0.5B CadQuery $-$ Text2CAD      & \multicolumn{2}{l}{F1 $+$0.294 \ci{$+$0.246, $+$0.341}} & \multicolumn{5}{l}{Dim $+$48.0 \ci{$+$41.5, $+$54.5}}
                                & \multicolumn{2}{c}{$+$14.8 \ci{$+$9.2, $+$20.6}} & \multicolumn{2}{c}{$+$20.9 \ci{$+$15.8, $+$25.8}} \\
0.5B command $-$ Text2CAD       & \multicolumn{2}{l}{F1 $+$0.312 \ci{$+$0.263, $+$0.360}} & \multicolumn{5}{l}{Dim $+$46.7 \ci{$+$39.8, $+$53.3}}
                                & \multicolumn{2}{c}{$-$2.2 \ci{$-$10.5, $+$5.5}} & \multicolumn{2}{c}{$+$8.6 \ci{$+$2.9, $+$14.0}} \\
0.5B CadQuery $-$ 0.5B command  & \multicolumn{7}{l}{(see Table~\ref{tab:main_results})}
                                & \multicolumn{2}{c}{$+$17.0 \ci{$+$8.6, $+$25.9}} & \multicolumn{2}{c}{$+$12.3 \ci{$+$7.2, $+$17.6}} \\
\bottomrule
\end{tabular}}
\end{table*}

\textbf{Text2CAD:} We re-evaluate the official Text2CAD checkpoint with the same intermediate-level prompts, greedy decoding and the official decoder, under exactly the protocol above. To avoid giving any model an unfair advantage, we use the 246 test items that also belong to the official Text2CAD test split, so that none of the compared models saw them during training. \begin{figure*}[!t]
  \centering
  \includegraphics[width=0.98\linewidth]{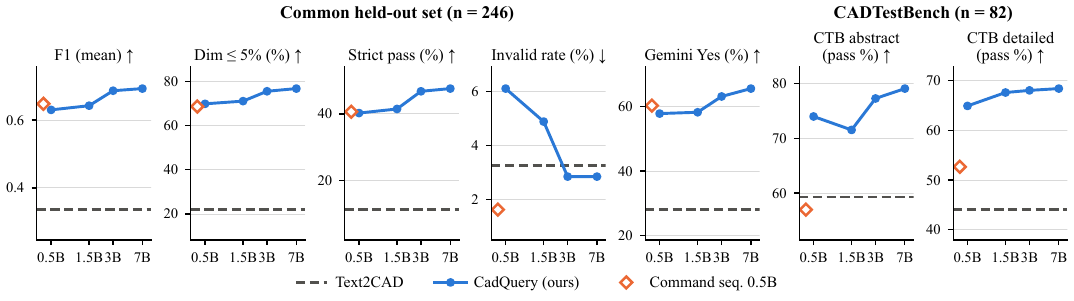}
  \vspace{-2mm}
  \caption{Comparison with Text2CAD on the common held-out set and on CADTestBench}
  \label{fig:external}
\end{figure*}

Table~\ref{tab:external} and Fig.~\ref{fig:external} summarize the comparison. The figure shows mean values, and the confidence intervals of the paired differences are given in the table. Table~\ref{tab:external}(a) shows that the fine-tuned Qwen2.5 models clearly outperform Text2CAD. For example, the 0.5B CadQuery model improves mean F1 from 0.336 to 0.630 ($+0.294$, $[+0.246, +0.341]$), Dim from 22.0\% to 69.9\%, and the Gemini judgment from 28.0\% to 57.7\%. \revise{The 0.5B command-sequence model, trained on the same backbone, achieves a similar improvement (F1 0.648, CD 12.89), which shows that fine-tuning a pretrained language model accounts for much of this gain. With CadQuery, the gain is obtained in a format the model already knows, and the output executes directly without dataset-specific conversion code. Sec.~\ref{sec:repr} compares the two representations under identical conditions.}

\textbf{CADTestBench:} To evaluate beyond our test split, we use CADTestBench~\cite{mallis2026text}, which pairs CAD parts from CADPrompt~\cite{alrashedy2025generating} with prompts at two levels of detail (abstract and detailed) and scores each output with a suite of executable tests on dimensions and ratios, geometry types, solid validity, spatial arrangement, topology and volume. Because its parts are drawn from the same broad CAD corpus as ours, we first remove every part that matches our training or validation data, as well as parts whose source we cannot verify, which leaves 82 parts. We report the mean fraction of tests passed per part, with invalid outputs scoring zero. The prompts differ in style and content from the Text2CAD descriptions used for training.

Table~\ref{tab:external}(b) shows a clear difference between the representations on this benchmark. With the same 0.5B backbone, CadQuery passes 17.0 points more tests than command sequences on abstract prompts ($[+8.6, +25.9]$) and 12.3 points more on detailed prompts ($[+7.2, +17.6]$). The command-sequence model fails to produce a valid solid for 21.95\% of the abstract prompts, whereas the CadQuery model produces a valid solid for all of them. CadQuery also exceeds Text2CAD by 14.8 and 20.9 points, and the 7B model passes the most tests on both prompt levels.

This result suggests that CadQuery transfers better than command sequences to prompts outside the training distribution, which is consistent with the model drawing on its pretrained knowledge of Python and CadQuery. \rtwo{However, this evidence is limited. The benchmark contains only 82 parts, and the inherent tendency of the pretrained model to map unfamiliar text to Python code may favor CadQuery, so the difference cannot be attributed to the representation alone.} We \rtwo{therefore} regard this result as supporting evidence \rtwo{ rather than a general conclusion about transferability}.

\subsection{Agreement Among Evaluators}\label{sec:evaluators}
\begin{table}[!t]
\vspace{-1mm}
\centering
\caption{Agreement among evaluators}
\label{tab:evaluator_agreement}
\renewcommand{\arraystretch}{1.15}
\setlength{\tabcolsep}{4pt}
\begin{tabular}{@{}lrr@{}}
\toprule
\textbf{Comparison} & \textbf{Agree.} & \textbf{Cohen's $\kappa$ [95\% CI]} \\
\midrule
\multicolumn{3}{@{}l}{\textit{Against human consensus, $n=159$}} \\
Gemini 3 Flash                          & 89.3\% & 0.783 [0.680, 0.874] \\
GPT-4.1                                 & 95.6\% & 0.907 [0.832, 0.969] \\
GPT-5.5                                 & 91.8\% & 0.834 [0.741, 0.911] \\
F1 $\ge 0.9$ (geometry)                 & 77.4\% & 0.549 [0.419, 0.672] \\
Strict engineering pass                 & 76.1\% & 0.543 [0.425, 0.657] \\
\midrule
\multicolumn{3}{@{}l}{\textit{Gemini vs.\ geometry, all valid predictions, $n=25{,}239$}} \\
vs.\ F1 $\ge 0.9$                       & 80.0\% & 0.596 \\
vs.\ strict engineering pass            & 73.5\% & 0.486 \\
\midrule
\multicolumn{3}{@{}l}{\textit{Gemini stability, $n=60$}} \\
Three identical calls, all agree        & 91.7\% & 0.86--0.93 \\
Shape-tolerant prompt                   & 91.7\% & 0.828 \\
Engineering-strict prompt               & 95.0\% & 0.899 \\
Viewpoint iso-2                         & 91.7\% & 0.831 \\
Viewpoint iso-3                         & 86.7\% & 0.731 \\
\bottomrule
\end{tabular}
\end{table}

Because the VLM judgment is a learned evaluator, we validate it against human judgment, against other VLMs and against deterministic geometry (Table~\ref{tab:evaluator_agreement}). Because our annotations are produced deterministically without any language model (Sec.~\ref{sec:annotation}), the annotation and the VLM-based evaluation are independent.

\textbf{Blinded human study:} We sample 180 prediction--reference pairs, 60 from each of three models (0.5B CadQuery, 1.5B CadQuery and 0.5B command sequences), balanced between Gemini Yes and No judgments. The two renders are shown in random left--right order, and the generating model and the Gemini judgment are hidden. Two human raters, both PhD students with CAD experience, independently judge for every pair whether it shows the same design, and two further VLMs, GPT-4.1 (\texttt{gpt-4.1-2025-04-14}) and GPT-5.5 (\texttt{gpt-5.5-2026-04-23}), judge the same pairs. The two humans agree on 88.3\% of the pairs (Cohen's $\kappa = 0.757$), and the three VLMs agree with each other to the same degree ($\kappa = 0.733$--$0.757$). On the 159 pairs where both humans agree, Gemini reaches $\kappa = 0.783$ with the human consensus, GPT-4.1 0.907 and GPT-5.5 0.834. VLM judgments thus agree with blinded human judgments as well as two humans agree with each other, and the agreement does not depend on a particular VLM. Fig.~\ref{fig:kappa} shows all pairwise agreements.

\begin{figure}[!t]
  \centering
  \includegraphics[width=0.95\linewidth]{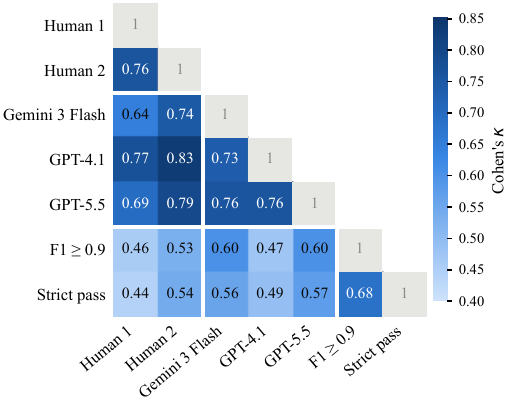}
  \vspace{-2mm}
  \caption{Pairwise agreement among evaluators}
  \label{fig:kappa}
\end{figure}

\textbf{Perceptual versus geometric judgment:} Fig.~\ref{fig:kappa} shows a clear block structure: humans and VLMs agree strongly with one another and moderately with deterministic geometry; the human consensus reaches $\kappa = 0.549$ with $\text{F1} \ge 0.9$ and 0.543 with the strict engineering pass. In the 27 pairs where Gemini and $\text{F1} \ge 0.9$ disagree but the human raters agree with one another, the humans side with Gemini in 23, so the VLM judgment closely reflects human visual assessment. Because both humans and VLMs judge rendered images, they also accept some predictions whose dimensions deviate in ways that only the raw-coordinate metrics detect. The same pattern holds at scale: on all 25,239 valid predictions of the three models used in the blinded study (0.5B CadQuery, 1.5B CadQuery and 0.5B command sequences), Gemini agrees with $\text{F1} \ge 0.9$ on 80.0\% of the items ($\kappa = 0.596$), answering Yes for 92.1\% of the items with $\text{F1} \ge 0.9$ and for 33.0\% of those below. The VLM judgment thus captures perceptual similarity as humans see it, while the deterministic metrics capture engineering accuracy. We report both and treat the deterministic metrics as primary.

\textbf{Stability:} On 60 blinded pairs, three identical Gemini calls give the same answer for 91.7\% of the pairs. Rewording the question in a more shape-tolerant or a more engineering-strict way changes the answer for 8.3\% and 5.0\% of the pairs ($\kappa = 0.828$ and 0.899 with the original question), and rendering from two other viewpoints changes it for 8.3\% and 13.3\% ($\kappa = 0.831$ and 0.731). Since the sample is balanced by the Gemini judgment, its Yes rates are not estimates of model accuracy.
\endgroup

\section{Discussion\label{sec:discussion}}

\subsection{Implications}
Our findings have several broader implications. First, generating executable CadQuery directly from natural language enables rapid prototyping: designers can describe concepts in plain text and immediately obtain parametric models, lowering the entry barriers for non-experts. Second, it can help novices learn the mapping between descriptions, CAD code, and 3D models. Third, CadQuery's lightweight, Python-native form is well suited for cloud-based CAD pipelines, where design repositories such as Onshape or Fusion 360 could offer natural language search and model generation.

\subsection{Limitations}
Several limitations remain. First, although our dataset of 170,000 text–CadQuery pairs is far larger than prior script-based datasets, it is derived from DeepCAD and thus reflects primarily simplified mechanical parts. Second, \revise{our size study covers four sizes of one model family up to 7B parameters, the largest size that fits our single-GPU training protocol.} Testing models at the 13B--70B scale would require more data diversity\revise{, quantity and compute}. \revise{Third, the representation comparison is made at 0.5B with a single command-sequence run, and the size study covers CadQuery only, so we do not claim that the improvement with size is specific to CadQuery.} \rtwo{Likewise, the external comparison on CADTestBench is limited to 82 parts.} Finally, \rtwo{because our training targets are obtained by deterministic translation, the generated scripts are structurally low-level: they reproduce the sketch-and-extrude steps of the source model with literal coordinates rather than named parameters or high-level features. This limits their semantic editability compared with human-written CadQuery code, and} our evaluation does not assess CAD usability aspects such as code modularity, parameterization, or ease of modification.

\subsection{Future Directions}
\revise{Several directions follow from this work. First, the training data can be extended in two ways: with more complex, industrially relevant parts and with multiple equivalent programs for the same geometry. Our annotations use one canonical program per part, whereas the same shape can be built with different sketches and operations. Training on such variants would teach the model the geometry rather than one coding style and make generation more robust. Second, because CadQuery is independent of the input modality, the same output format can be generated from images, sketches, or engineering drawings by replacing the backbone with a vision-language model, as explored for images in concurrent work~\cite{doris2026cad}. Third, because CadQuery programs are directly executable, our dataset and fine-tuned models provide a starting point for reinforcement learning with execution- and geometry-based rewards, as explored in concurrent work~\cite{guan2026cad}.}

\section{Conclusion}\label{sec:conclusion}
We introduce a large-scale dataset of 170,000 text–CadQuery pairs\revise{, each verified against its source CAD model,} and study Text-to-CadQuery as a scalable approach to CAD generation. Our method directly generates executable CadQuery code from natural language, replacing task-specific command sequences used in prior work. This shift \revise{yields readable and editable programs that execute directly without dataset-specific conversion code. Fine-tuned from a pretrained backbone, our models substantially outperform the Text2CAD baseline.} \revise{Compared with a command-sequence baseline on the same backbone, CadQuery reaches the same shape accuracy, is more accurate in absolute dimensions and transfers better \rtwo{on a small external benchmark}, consistent with the model drawing on its pretrained knowledge of Python.} More importantly, our approach enables seamless integration with future advances in foundation models, such as enhanced multimodal understanding, spatial reasoning, and code generation capabilities, without requiring changes to the training procedure or output format. As model size increases, we observe a clear scaling trend in accuracy and executability\revise{, with the 7B model best on every metric}. These findings validate our framework and open new directions for extending CadQuery-based generation to diverse inputs and larger models.

\appendices

\section{Background: CAD Command Sequence Representation\label{supplementary:command_sequence_background}}
Several prior works, including DeepCAD~\cite{wu2021deepcad} and Text2CAD~\cite{khan2024text2cad}, represent a CAD model as a fixed-length sequence of parametric commands that mimics the modeling history in CAD software. Six command types are used (\texttt{Line}, \texttt{Arc}, \texttt{Circle}, \texttt{SOL}, \texttt{Extrude}, and \texttt{EOS}), each paired with up to 16 quantized parameters (e.g., coordinates, radius, extrusion depth, and boolean type). Each command is embedded as a one-hot command type together with one-hot parameter vectors and a positional encoding indicating its index in the sequence. The sequence is padded or truncated to a fixed length (typically 60) to support batch training and decoding.

\paragraph{Example}
\begingroup
\rtwo{Text2CAD~\cite{khan2024text2cad} encodes these commands as a sequence of integer token pairs, its CAD vector, from which our CadQuery annotations are derived and which our command-sequence baseline generates. A point on a curve uses both numbers as its $(x, y)$ coordinates. Every other token carries a single number followed by a filler 0. Values 0--6 are structural markers (padding, start or end of the sequence, and the end of a sketch, face, loop, curve or extrusion), 7--10 give the Boolean operation (new body, join, cut, intersect), and 11--266 are parameters quantized to 256 levels (value $-\,11 \in \{0, \dots, 255\}$), so that 139 is the middle level. The table below lists only the first construction step of a hollow-cylinder example from our test set and the complete sequence contains 36 token pairs. The second step (not shown) builds a smaller cylinder and subtracts it from the first (cut token \texttt{[9,0]}), and the sequence ends with \texttt{[1,0]}. The same model is shown as CadQuery code in Appendix~\ref{supplementary: dataset_examples}.}

\begin{center}
\scriptsize
\setlength{\tabcolsep}{3pt}
\begin{tabularx}{\linewidth}{@{}>{\ttfamily}p{4.1cm}X@{}}
\textbf{\rmfamily Token pairs} & \textbf{Meaning} \\
\hline
{}[1,0] & start of sequence \\
{}[139,139] [139,266] & circle: center and a point on the circle \\
{}[5,0] [4,0] [3,0] [2,0] & end of curve, loop, face and sketch \\
{}[177,0] [139,0] & extrusion extents (two directions) \\
{}[139,0] [139,0] [139,0] & sketch-plane origin $(x, y, z)$ \\
{}[139,0] [139,0] [139,0] & sketch-plane orientation $(\theta, \phi, \gamma)$ \\
{}[7,0] [107,0] [6,0] & new body, sketch scale, end of extrusion \\
\hline
\end{tabularx}
\end{center}
\endgroup

\rtwo{After generation, such a sequence must be converted by dataset-specific utilities (e.g., \texttt{save\_stp()} or \texttt{save\_points()}) to obtain a STEP or mesh file or a point cloud.}

\section{Dataset Examples\label{supplementary: dataset_examples}}
\begingroup
\paragraph{\textbf{Text Description}}
Start by creating a solid cylindrical object with a flat circular top and bottom. The cylinder has a length and width of 0.75 and a height of 0.3. Next, construct a hollow center within this cylinder. The hollow section has a length and width of 0.42 and a height of 0.3. These two parts are aligned such that the hollow center is centered within the larger cylinder, creating a hollow cylindrical object.
\paragraph{\textbf{CadQuery}}
\begin{lstlisting}
import cadquery as cq

# Part 1: new
part_1 = (
    cq.Workplane("XY")
    .moveTo(0.3765, 0.3765)
    .circle(0.3735)
    .extrude(0.2969)
)

# Part 2: cut
part_2 = (
    cq.Workplane("XY")
    .moveTo(0.2118, 0.2118)
    .circle(0.2101)
    .extrude(0.2969)
)
part_2 = part_2.translate((0.164062, 0.164062, 0))

result = part_1
result = result.cut(part_2)

cq.exporters.export(result, "./00099764.stl")
\end{lstlisting}
This is the same model as the CAD vector in Appendix~\ref{supplementary:command_sequence_background}. The heights (0.2969 rather than the stated 0.3) reproduce the quantized values of the CAD vector exactly.
\endgroup

\section{Comparison and Examples of Parametric CAD Tools\label{supplementary: parametric_cad_example}}

To support CAD model generation, we compare several open-source parametric modeling tools. Table~\ref{tab:tool_comparison} summarizes their scripting capabilities and integration potential with Python-based pipelines. CadQuery offers the best trade-off between usability and flexibility, making it well-suited for programmatic shape synthesis.

\begin{table}[!h]
\vspace{-1mm}
\centering
\caption{Comparison of Parametric CAD Tools\label{tab:tool_comparison}}
\renewcommand{\arraystretch}{1.2}
\setlength{\tabcolsep}{3pt}
\footnotesize
\begin{tabular}{@{}lccc>{\raggedright\arraybackslash}p{2.3cm}@{}}
\toprule
\textbf{Tool} & \textbf{Software} & \textbf{Language} & \textbf{Difficulty} & \textbf{Python integration} \\
\midrule
FreeCAD     & Yes & Python       & Low    & Limited (GUI-bound API) \\
OpenSCAD    & Yes & OpenSCAD DSL & Medium & None \\
PythonOCC   & No  & Python       & High   & Low-level, verbose \\
\textbf{CadQuery} & \textbf{No} & \textbf{Python} & \textbf{Low} & \textbf{Excellent (Jupyter, scripting)} \\
\bottomrule
\end{tabular}
\end{table}

\noindent\textbf{Modeling Examples:}

\noindent\textbf{FreeCAD}
\begin{lstlisting}
import FreeCAD, Part
doc = FreeCAD.newDocument("Example")
box = Part.makeBox(10, 10, 10)
Part.show(box)
doc.recompute()
\end{lstlisting}

\noindent\textbf{OpenSCAD}
\begin{lstlisting}
cube([10, 10, 10]);
\end{lstlisting}

\noindent\textbf{CadQuery}
\begin{lstlisting}
import cadquery as cq
result = cq.Workplane("XY").box(10, 10, 10)
cq.exporters.export(result, 'box.stl')    
\end{lstlisting}

\noindent\textbf{PythonOCC}
\begin{lstlisting}
from OCC.Core.BRepPrimAPI import BRepPrimAPI_MakeBox
from OCC.Display.SimpleGui import init_display

box = BRepPrimAPI_MakeBox(10, 10, 10).Shape()
display, start_display, _, _ = init_display()
display.DisplayShape(box, update=True)
start_display()
\end{lstlisting}

\bibliographystyle{IEEEtran}
\bibliography{references}

@article{xie2025transformer,
  title={Transformer-Based Offline Printing Strategy Design for Large Format Additive Manufacturing},
  author={Xie, Haoyang and Hoskins, Dylan and Rowe, Kyle and Ju, Feng},
  journal={Journal of Computing and Information Science in Engineering},
  volume={25},
  number={2},
  year={2025},
  publisher={American Society of Mechanical Engineers Digital Collection}
}

@article{fathizadan2022dynamic,
  title={Dynamic material deposition control for large-scale additive manufacturing},
  author={Fathizadan, Sepehr and Ju, Feng and Wang, Feifan and Rowe, Kyle and Hofmann, Nils},
  journal={IISE Transactions},
  volume={54},
  number={9},
  pages={817--831},
  year={2022},
  publisher={Taylor \& Francis}
}

@article{xie2026agent,
  title={Agent-Based Framework for Engineering Document Comprehension With Large Language Models},
  author={Xie, Haoyang and Ju, Feng},
  journal={Journal of Computing and Information Science in Engineering},
  volume={26},
  number={7},
  pages={071003},
  year={2026},
  publisher={American Society of Mechanical Engineers}
}

@inproceedings{koch2019abc,
  title={{ABC}: A big {CAD} model dataset for geometric deep learning},
  author={Koch, Sebastian and Matveev, Albert and Jiang, Zhongshi and Williams, Francis and Artemov, Alexey and Burnaev, Evgeny and Alexa, Marc and Zorin, Denis and Panozzo, Daniele},
  booktitle={Proceedings of the IEEE/CVF conference on computer vision and pattern recognition},
  pages={9601--9611},
  year={2019}
}

@article{seff2020sketchgraphs,
  title={{SketchGraphs}: A large-scale dataset for modeling relational geometry in computer-aided design},
  author={Seff, Ari and Ovadia, Yaniv and Zhou, Wenda and Adams, Ryan P},
  journal={arXiv preprint arXiv:2007.08506},
  year={2020}
}

@inproceedings{wu2021deepcad,
  title={{DeepCAD}: A deep generative network for computer-aided design models},
  author={Wu, Rundi and Xiao, Chang and Zheng, Changxi},
  booktitle={Proceedings of the IEEE/CVF International Conference on Computer Vision},
  pages={6772--6782},
  year={2021}
}

@article{willis2021fusion,
  title={{Fusion 360 Gallery}: A dataset and environment for programmatic {CAD} construction from human design sequences},
  author={Willis, Karl DD and Pu, Yewen and Luo, Jieliang and Chu, Hang and Du, Tao and Lambourne, Joseph G and Solar-Lezama, Armando and Matusik, Wojciech},
  journal={ACM Transactions on Graphics (TOG)},
  volume={40},
  number={4},
  pages={1--24},
  year={2021},
  publisher={ACM New York, NY, USA}
}

@inproceedings{dupont2022cadops,
  title={{CADOps-Net}: Jointly learning {CAD} operation types and steps from boundary-representations},
  author={Dupont, Elona and Cherenkova, Kseniya and Kacem, Anis and Ali, Sk Aziz and Arzhannikov, Ilya and Gusev, Gleb and Aouada, Djamila},
  booktitle={2022 International Conference on 3D Vision (3DV)},
  pages={114--123},
  year={2022},
  organization={IEEE}
}

@inproceedings{xu2022skexgen,
  title={{SkexGen}: Autoregressive generation of {CAD} construction sequences with disentangled codebooks},
  author={Xu, Xiang and Willis, Karl DD and Lambourne, Joseph G and Cheng, Chin-Yi and Jayaraman, Pradeep Kumar and Furukawa, Yasutaka},
  booktitle={International Conference on Machine Learning},
  pages={24698--24724},
  year={2022},
  organization={PMLR}
}

@inproceedings{zhou2023cadparser,
  title={{CADParser}: A learning approach of sequence modeling for {B-Rep} {CAD}},
  author={Zhou, Shengdi and Tang, Tianyi and Zhou, Bin},
  booktitle={IJCAI},
  pages={1804--1812},
  year={2023}
}

@article{jayaraman2022solidgen,
  title={{SolidGen}: An autoregressive model for direct {B-rep} synthesis},
  author={Jayaraman, Pradeep Kumar and Lambourne, Joseph G and Desai, Nishkrit and Willis, Karl DD and Sanghi, Aditya and Morris, Nigel JW},
  journal={arXiv preprint arXiv:2203.13944},
  year={2022}
}

@article{xu2024brepgen,
  title={{BrepGen}: A {B-rep} generative diffusion model with structured latent geometry},
  author={Xu, Xiang and Lambourne, Joseph and Jayaraman, Pradeep and Wang, Zhengqing and Willis, Karl and Furukawa, Yasutaka},
  journal={ACM Transactions on Graphics (TOG)},
  volume={43},
  number={4},
  pages={1--14},
  year={2024},
  publisher={ACM New York, NY, USA}
}

@inproceedings{you2025img2cad,
  title={{Img2CAD}: Reverse engineering {3D} {CAD} models from images through {VLM}-assisted conditional factorization},
  author={You, Yang and Uy, Mikaela Angelina and Han, Jiaqi and Thomas, Rahul and Zhang, Haotong and Du, Yi and Chen, Hansheng and Engelmann, Francis and You, Suya and Guibas, Leonidas},
  booktitle={Proceedings of the SIGGRAPH Asia 2025 Conference Papers},
  pages={1--12},
  year={2025}
}

@article{yuan2024openecad,
  title={{OpenECAD}: An efficient visual language model for editable {3D-CAD} design},
  author={Yuan, Zhe and Shi, Jianqi and Huang, Yanhong},
  journal={Computers \& Graphics},
  volume={124},
  pages={104048},
  year={2024},
  publisher={Elsevier}
}

@article{chen2025img2cad,
  title={{Img2CAD}: Conditioned {3-D} {CAD} model generation from single image with structured visual geometry},
  author={Chen, Tianrun and Yu, Chunan and Hu, Yuanqi and Li, Jing and Xu, Tao and Cao, Runlong and Zhu, Lanyun and Zang, Ying and Zhang, Yong and Li, Zejian and others},
  journal={IEEE Transactions on Industrial Informatics},
  volume={21},
  number={11},
  pages={8539--8549},
  year={2025},
  publisher={IEEE}
}

@article{badagabettu2024query2cad,
  title={{Query2CAD}: Generating {CAD} models using natural language queries},
  author={Badagabettu, Akshay and Yarlagadda, Sai Sravan and Farimani, Amir Barati},
  journal={arXiv preprint arXiv:2406.00144},
  year={2024}
}

@article{khan2024text2cad,
  title={{Text2CAD}: Generating sequential {CAD} designs from beginner-to-expert level text prompts},
  author={Khan, Mohammad S and Sinha, Sankalp and Sheikh, Talha U and Stricker, Didier and Ali, Sk A and Afzal, Muhammad Z},
  journal={Advances in Neural Information Processing Systems},
  volume={37},
  pages={7552--7579},
  year={2024}
}

@article{deitke2023objaverse,
  title={{Objaverse-XL}: A universe of {10M+} {3D} objects},
  author={Deitke, Matt and Liu, Ruoshi and Wallingford, Matthew and Ngo, Huong and Michel, Oscar and Kusupati, Aditya and Fan, Alan and Laforte, Christian and Voleti, Vikram and Gadre, Samir Yitzhak and others},
  journal={Advances in Neural Information Processing Systems},
  volume={36},
  pages={35799--35813},
  year={2023}
}

@article{xu2024cad,
  title={{CAD-MLLM}: Unifying multimodality-conditioned {CAD} generation with {MLLM}},
  author={Xu, Jingwei and Zhao, Zibo and Wang, Chenyu and Liu, Wen and Ma, Yi and Gao, Shenghua},
  journal={arXiv preprint arXiv:2411.04954},
  year={2024}
}

@article{li2022free2cad,
  title={{Free2CAD}: Parsing freehand drawings into {CAD} commands},
  author={Li, Changjian and Pan, Hao and Bousseau, Adrien and Mitra, Niloy J},
  journal={ACM Transactions on Graphics (TOG)},
  volume={41},
  number={4},
  pages={1--16},
  year={2022},
  publisher={ACM New York, NY, USA}
}

@inproceedings{liu2024point2cad,
  title={{Point2CAD}: Reverse engineering {CAD} models from {3D} point clouds},
  author={Liu, Yujia and Obukhov, Anton and Wegner, Jan Dirk and Schindler, Konrad},
  booktitle={Proceedings of the IEEE/CVF Conference on Computer Vision and Pattern Recognition},
  pages={3763--3772},
  year={2024}
}

@article{wang2025point2skh,
  title={{Point2Skh}: End-to-end parametric primitive inference from point clouds with improved denoising transformer},
  author={Wang, Cheng and Sun, Wenyu and Ma, Xinzhu and Deng, Fei},
  journal={Computer-Aided Design},
  volume={181},
  pages={103838},
  year={2025},
  publisher={Elsevier}
}

@article{sun2025large,
  title={Large language models for computer-aided design ({LLM4CAD}) fine-tuned: Dataset and experiments},
  author={Sun, Yuewan and Li, Xingang and Sha, Zhenghui},
  journal={Journal of Mechanical Design},
  pages={1--19},
  year={2025}
}

@article{hunde2022future,
  title={Future prospects of computer-aided design ({CAD})--{A} review from the perspective of artificial intelligence ({AI}), extended reality, and {3D} printing},
  author={Hunde, Bonsa Regassa and Woldeyohannes, Abraham Debebe},
  journal={Results in Engineering},
  volume={14},
  pages={100478},
  year={2022},
  publisher={Elsevier}
}

@article{kaplan2020scaling,
  title={Scaling laws for neural language models},
  author={Kaplan, Jared and McCandlish, Sam and Henighan, Tom and Brown, Tom B and Chess, Benjamin and Child, Rewon and Gray, Scott and Radford, Alec and Wu, Jeffrey and Amodei, Dario},
  journal={arXiv preprint arXiv:2001.08361},
  year={2020}
}

@article{riegel2016freecad,
  title={{FreeCAD}},
  author={Riegel, Juergen and Mayer, Werner and van Havre, Yorik},
  journal={Freecadspec2002. pdf},
  year={2016}
}

@article{machado2019parametric,
  title={Parametric {CAD} modeling for open source scientific hardware: Comparing {OpenSCAD} and {FreeCAD} {Python} scripts},
  author={Machado, Felipe and Malpica, Norberto and Borromeo, Susana},
  journal={Plos one},
  volume={14},
  number={12},
  pages={e0225795},
  year={2019},
  publisher={Public Library of Science San Francisco, CA USA}
}

@software{cadquery_2024,
  title        = {{CadQuery/cadquery: CadQuery 2.4.0}},
  author       = {Jeremy Wright and thebluedirt and Marcus Boyd and Lorenz and Innovations Technology Solutions and Hasan Yavuz ÖZDERYA and Bruno Agostini and Jojain and Michael Greminger and Seth Fischer and Justin Buchanan and cactrot and huskier and Ruben and Iulian Onofrei and Miguel Sánchez de León Peque and Martin Budden and Hecatron and Peter Boin and Wink Saville and Pavel M. Penev and Bryan Weissinger and M. Greyson Christoforo and Jack Case and AGD and Paul Jurczak and nopria and moeb and jdegenstein},
  year         = 2024,
  month        = jan,
  publisher    = {Zenodo},
  version      = {2.4.0},
  doi          = {10.5281/zenodo.10513848},
  url          = {https://doi.org/10.5281/zenodo.10513848}
}

@article{yang2024qwen2,
  title={{Qwen2.5} technical report},
  author={Yang, An and Yang, Baosong and Zhang, Beichen and Hui, Binyuan and Zheng, Bo and Yu, Bowen and Li, Chengyuan and Liu, Dayiheng and Huang, Fei and Wei, Haoran and others},
  journal={arXiv preprint arXiv:2412.15115},
  year={2024}
}

@article{li2025llm4cad,
  title={{LLM4CAD}: Multimodal large language models for three-dimensional computer-aided design generation},
  author={Li, Xingang and Sun, Yuewan and Sha, Zhenghui},
  journal={Journal of Computing and Information Science in Engineering},
  volume={25},
  number={2},
  year={2025},
  publisher={American Society of Mechanical Engineers Digital Collection}
}

@article{team2023gemini,
  title={{Gemini}: A family of highly capable multimodal models},
  author={Team, Gemini and Anil, Rohan and Borgeaud, Sebastian and Alayrac, Jean-Baptiste and Yu, Jiahui and Soricut, Radu and Schalkwyk, Johan and Dai, Andrew M and Hauth, Anja and Millican, Katie and others},
  journal={arXiv preprint arXiv:2312.11805},
  year={2023}
}

@article{achiam2023gpt,
  title={{GPT-4} technical report},
  author={Achiam, Josh and Adler, Steven and Agarwal, Sandhini and Ahmad, Lama and Akkaya, Ilge and Aleman, Florencia Leoni and Almeida, Diogo and Altenschmidt, Janko and Altman, Sam and Anadkat, Shyamal and others},
  journal={arXiv preprint arXiv:2303.08774},
  year={2023}
}

@inproceedings{wu2024gpt,
  title={{GPT-4V(ision)} is a human-aligned evaluator for text-to-{3D} generation},
  author={Wu, Tong and Yang, Guandao and Li, Zhibing and Zhang, Kai and Liu, Ziwei and Guibas, Leonidas and Lin, Dahua and Wetzstein, Gordon},
  booktitle={Proceedings of the IEEE/CVF conference on computer vision and pattern recognition},
  pages={22227--22238},
  year={2024}
}

@misc{blender,
  title = {{Blender} -- a {3D} modelling and rendering package},
  author = {{Blender Online Community}},
  year = {2024},
  url = {https://www.blender.org},
  note = {Version 4.1}
}

@inproceedings{fan2017point,
  title={A point set generation network for {3D} object reconstruction from a single image},
  author={Fan, Haoqiang and Su, Hao and Guibas, Leonidas J},
  booktitle={Proceedings of the IEEE conference on computer vision and pattern recognition},
  pages={605--613},
  year={2017}
}

@inproceedings{wu20153d,
  title={{3D ShapeNets}: A deep representation for volumetric shapes},
  author={Wu, Zhirong and Song, Shuran and Khosla, Aditya and Yu, Fisher and Zhang, Linguang and Tang, Xiaoou and Xiao, Jianxiong},
  booktitle={Proceedings of the IEEE conference on computer vision and pattern recognition},
  pages={1912--1920},
  year={2015}
}

@inproceedings{devlin2019bert,
  title={{BERT}: Pre-training of deep bidirectional transformers for language understanding},
  author={Devlin, Jacob and Chang, Ming-Wei and Lee, Kenton and Toutanova, Kristina},
  booktitle={Proceedings of the 2019 conference of the North American chapter of the association for computational linguistics: human language technologies, volume 1 (long and short papers)},
  pages={4171--4186},
  year={2019}
}

@misc{zooZooModel,
	author = {},
	title = {{Z}oo: {A}{I} {C}{A}{D} {M}odel {G}enerator --- zoo.dev},
	howpublished = {\url{https://zoo.dev/text-to-cad}},
	year = {},
	note = {[Accessed 02-05-2025]},
}

@article{hui2024qwen2,
  title={{Qwen2.5-Coder} technical report},
  author={Hui, Binyuan and Yang, Jian and Cui, Zeyu and Yang, Jiaxi and Liu, Dayiheng and Zhang, Lei and Liu, Tianyu and Zhang, Jiajun and Yu, Bowen and Lu, Keming and others},
  journal={arXiv preprint arXiv:2409.12186},
  year={2024}
}

@article{doris2026cad,
  title={{CAD-Coder}: An open-source vision-language model for computer-aided design code generation},
  author={Doris, Anna C and Alam, Ferdous and Heyrani Nobari, Amin and Ahmed, Faez},
  journal={Journal of Mechanical Design},
  volume={148},
  number={7},
  pages={071702},
  year={2026},
  publisher={American Society of Mechanical Engineers}
}

@article{guan2026cad,
  title={{CAD-Coder}: Text-to-{CAD} generation with chain-of-thought and geometric reward},
  author={Guan, Yandong and Wang, Xilin and Xing, Ximing and Zhang, Jing and Xu, Dong and Yu, Qian},
  journal={Advances in Neural Information Processing Systems},
  volume={38},
  pages={59765--59789},
  year={2026}
}

@article{qin2026brep,
  title={{BRep-GD}: A graph diffusion model for {CAD} boundary representation generation},
  author={Qin, Feiwei and Luo, Chenqi and Hou, Junhao and Fang, Meie and Liu, Ligang},
  journal={IEEE Transactions on Visualization and Computer Graphics},
  year={2026},
  publisher={IEEE}
}

@inproceedings{hou2026hifi,
  title={{HiFi-BRep}: High-fidelity latent representation for robust {B-Rep} generation},
  author={Hou, Junhao and Luo, Chenqi and Wang, Pufan and Lu, Jiaying and Liu, Yusheng and Qin, Feiwei and Fang, Meie and Zhou, Kun},
  booktitle={Proceedings of the IEEE/CVF Conference on Computer Vision and Pattern Recognition},
  pages={27199--27208},
  year={2026}
}

@inproceedings{qin2025drawing2cad,
  title={{Drawing2CAD}: Sequence-to-sequence learning for {CAD} generation from vector drawings},
  author={Qin, Feiwei and Lu, Shichao and Hou, Junhao and Wang, Changmiao and Fang, Meie and Liu, Ligang},
  booktitle={Proceedings of the 33rd ACM international conference on multimedia},
  pages={10573--10582},
  year={2025}
}

@article{mallis2026text,
  title={{Text-to-CAD} evaluation with {CADTests}},
  author={Mallis, Dimitrios and Wang, Marco and Karadeniz, Ahmet Serdar and Ricci, Elisa and Kacem, Anis and Aouada, Djamila},
  journal={arXiv preprint arXiv:2605.07807},
  year={2026}
}

@inproceedings{alrashedy2025generating,
  title={Generating {CAD} code with vision-language models for {3D} designs},
  author={Alrashedy, Kamel and Tambwekar, Pradyumna and Zaidi, Zulfiqar Haider and Langwasser, Megan and Xu, Wei and Gombolay, Matthew},
  booktitle={International Conference on Learning Representations},
  volume={2025},
  pages={52236--52262},
  year={2025}
}

\makeatletter
\def\@IEEEBIOskipN{1.5\baselineskip}
\makeatother
\begin{IEEEbiography}[{\includegraphics[width=1in,height=1.25in,keepaspectratio]{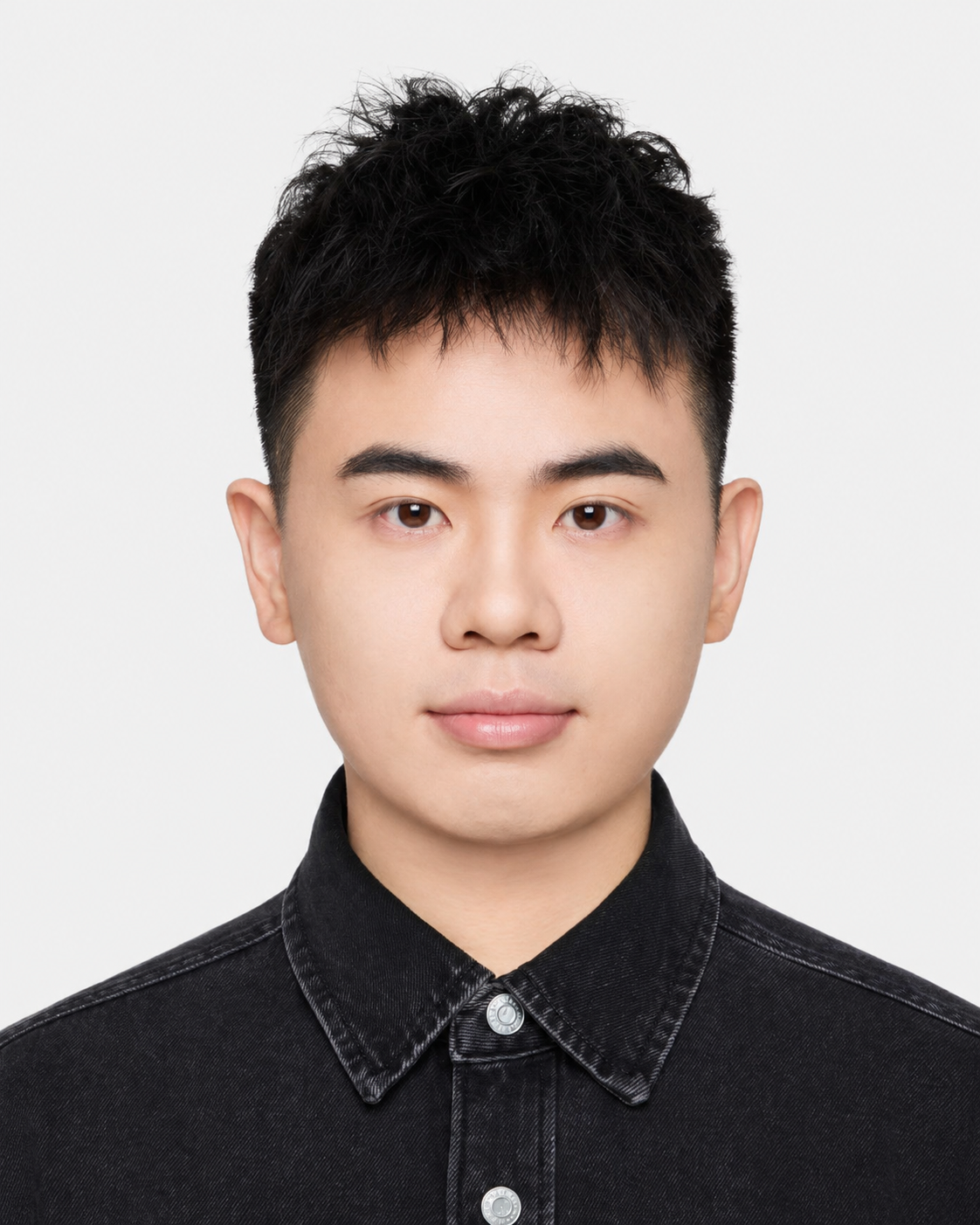}}]{Haoyang Xie}
received his Bachelor's degree in Materials Science from Lanzhou University of Technology, Lanzhou, China, in 2020. Continuing his academic journey, he pursued a Master's degree in Industrial Engineering from Rice University, Houston, USA, graduating in 2022. He is currently a PhD student in Industrial Engineering at Arizona State University, Tempe, USA. His research interests include artificial intelligence and its applications in large format additive manufacturing.
\end{IEEEbiography}

\begin{IEEEbiography}[{\includegraphics[width=1in,height=1.25in,keepaspectratio]{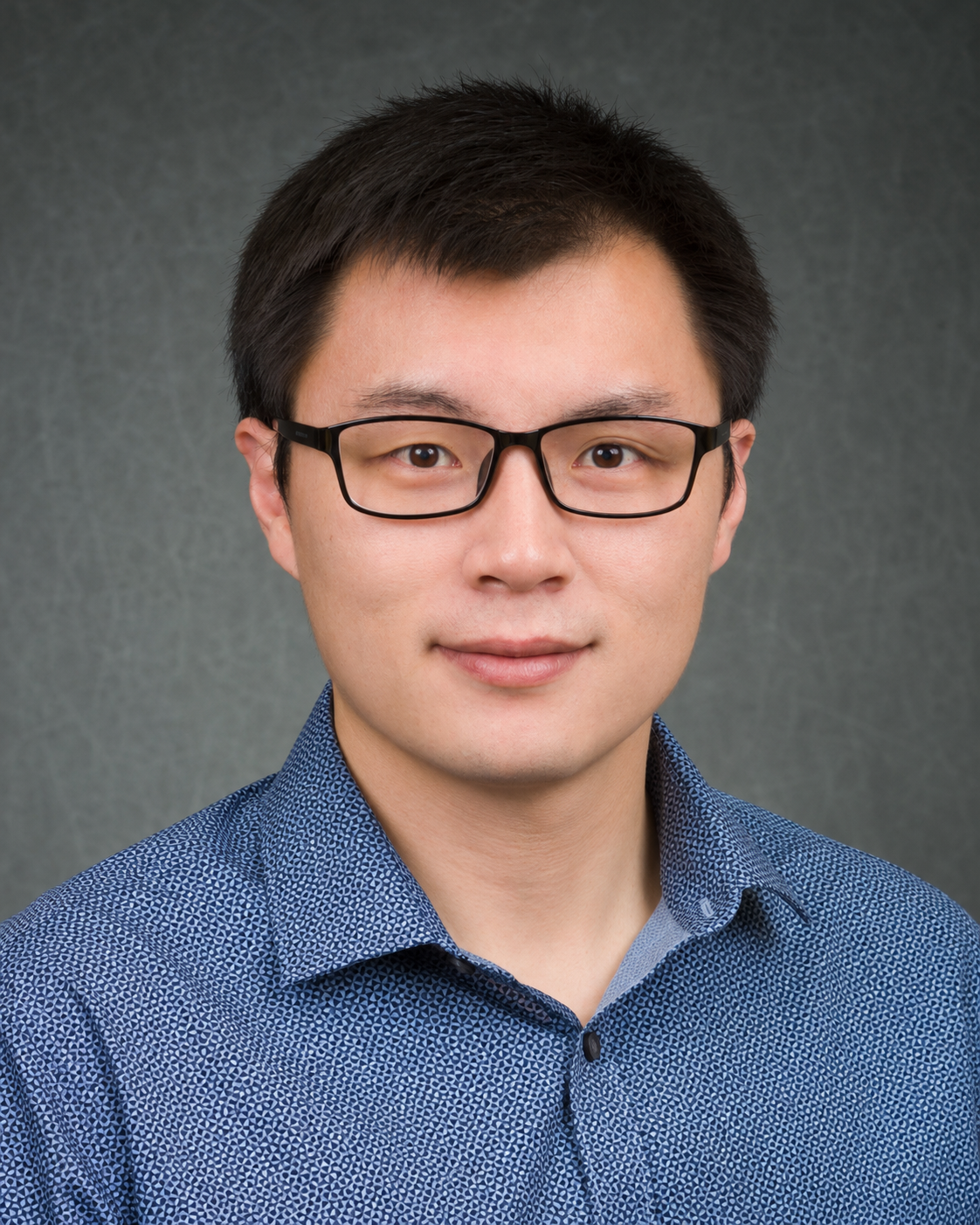}}]{Feng Ju}
\textit{(Member, IEEE)} received the B.S. degree from Shanghai Jiao Tong University, Shanghai, China, in 2010, and the M.S. degree in electrical and computer engineering and the Ph.D. degree in industrial and systems engineering from the University of Wisconsin--Madison, Madison, WI, USA, in 2011 and 2015, respectively. He is currently the Avnet Professor of Supply Chain Networks in Industrial Engineering with the School of Computing and Augmented Intelligence, Arizona State University, Tempe, AZ, USA. His research interests include AI, machine learning, and optimization for additive manufacturing, manufacturing systems and supply chain. He is a member of the Institute of Operations Research and the Management Sciences, the Institute of Industrial and Systems Engineers, the Institute of Electrical and Electronics Engineers, and the American Society of Mechanical Engineers. He has been a recipient of multiple awards, including the Best Paper Award in IISE Transactions, IFAC MIM, and IEEE CASE.
\end{IEEEbiography}
\vfill

\end{document}